\documentclass[11pt]{article}

\newif\ifabstractonly
\abstractonlyfalse

\usepackage[preprint]{acl}

\usepackage{times}
\usepackage{latexsym}

\usepackage[T1]{fontenc}

\usepackage[utf8]{inputenc}

\usepackage{microtype}

\usepackage{inconsolata}

\usepackage{graphicx}

\usepackage{amsmath}
\usepackage{amsfonts}
\usepackage{booktabs}
\usepackage[capitalize,noabbrev]{cleveref}
\usepackage{enumitem}
\usepackage[bb=pazo]{mathalpha}
\usepackage{tikz}
\usepackage{adjustbox}

\newcommand{\myParagraphWithoutTitle}{\paragraph{}\hspace{-1em}}

\newcommand{\myDisplaySkip}{0.4em}
\AtBeginDocument{
  \setlength{\abovedisplayskip}{\myDisplaySkip}
  \setlength{\belowdisplayskip}{\myDisplaySkip}
  \setlength{\abovedisplayshortskip}{\myDisplaySkip}
  \setlength{\belowdisplayshortskip}{\myDisplaySkip}
}
\allowdisplaybreaks

\NewDocumentEnvironment{myEquation}{+b}{\begin{equation*}\adjustbox{max width=\linewidth}{$#1$}\end{equation*}}{}

\newcommand{\declareGenericFig}[5]{
  \expandafter\newcommand\csname myFloat@#1\endcsname[2]{
    \begin{#4}[##1]
      \centering
      \includegraphics[width=#5\dimexpr ##2\linewidth \relax]{figures/#1}
      \caption[#2]{#2. #3}
      \label{fig:#1}
    \end{#4}
  }
}
\newcommand{\declareFig}[3]{\declareGenericFig{#1}{#2}{#3}{figure}{1}}
\newcommand{\declareWideFig}[3]{\declareGenericFig{#1}{#2}{#3}{figure*}{1}}

\definecolor{myRoundedFigBorderColor}{HTML}{666666}
\newcommand{\declareRoundedGenericFig}[6]{
  \expandafter\newcommand\csname myFloat@#1\endcsname[2]{
    \begin{#5}[##1]
      \centering
      \begin{tikzpicture}
        \node[
          draw=myRoundedFigBorderColor,
          line width=1pt,
          rounded corners=5pt,
          inner sep=#2
        ] {\includegraphics[width=#6\dimexpr ##2\linewidth \relax]{figures/#1}};
      \end{tikzpicture}
      \caption[#3]{#3. #4}
      \label{fig:#1}
    \end{#5}
  }
}
\newcommand{\declareRoundedFig}[4]{\declareRoundedGenericFig{#1}{#2}{#3}{#4}{figure}{1}}
\newcommand{\declareRoundedWideFig}[4]{\declareRoundedGenericFig{#1}{#2}{#3}{#4}{figure*}{1}}

\newcommand{\declareGenericTab}[6]{
  \expandafter\newcommand\csname myFloat@#1\endcsname[2]{
    \begin{#5}[##1]
      \centering
      \caption[#2]{#2. #3}
      \label{tab:#1}
      \resizebox{#6\dimexpr ##2\linewidth \relax}{!}{#4}
    \end{#5}
  }
}
\newcommand{\declareTab}[4]{\declareGenericTab{#1}{#2}{#3}{#4}{table}{1}}

\newcommand{\myFloat}[3]{\csname myFloat@#1\endcsname{#2}{#3}}
\declareWideFig{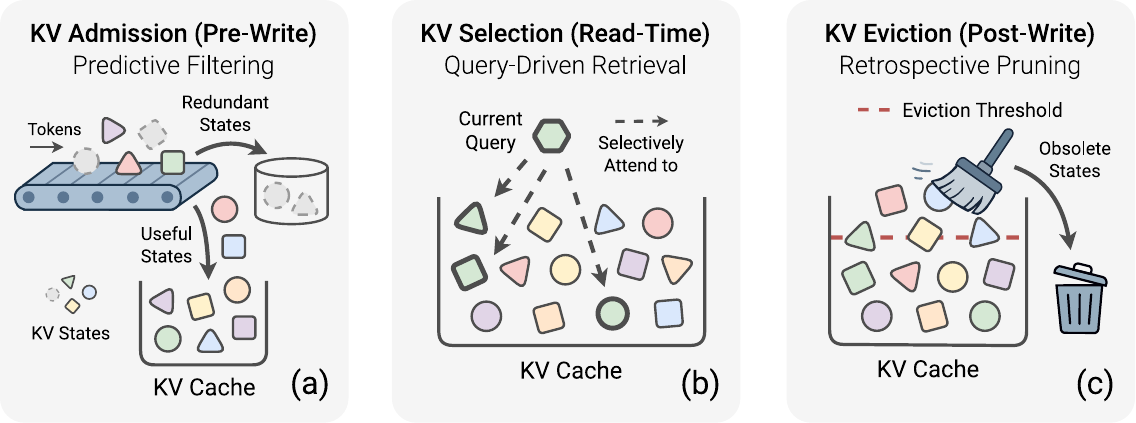}{A unified framework for KV management}{(a) \textit{KV Admission (Ours)} predictively filters redundant tokens pre-write. (b) \textit{KV Selection} selectively attends to relevant tokens at read-time. (c) \textit{KV Eviction} retrospectively prunes obsolete tokens post-write.}

\declareFig{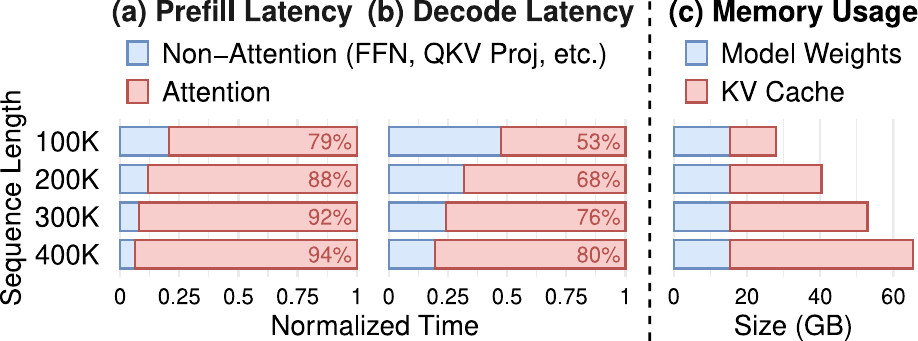}{Attention bottleneck in long-context LLM inference}{Measured on Llama-3.1-8B on an H200 GPU.}

\declareFig{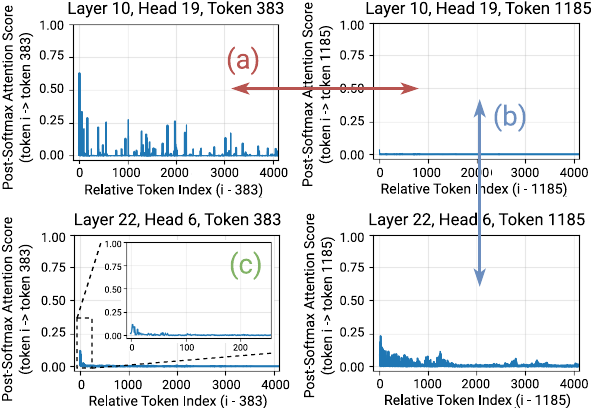}{Heterogeneity of token utility}{}

\declareFig{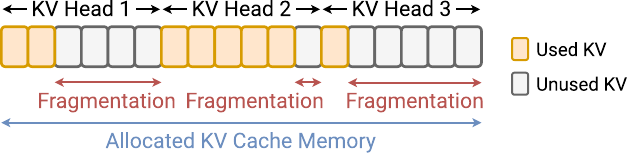}{Memory fragmentation with ragged KV states}{}

\declareFig{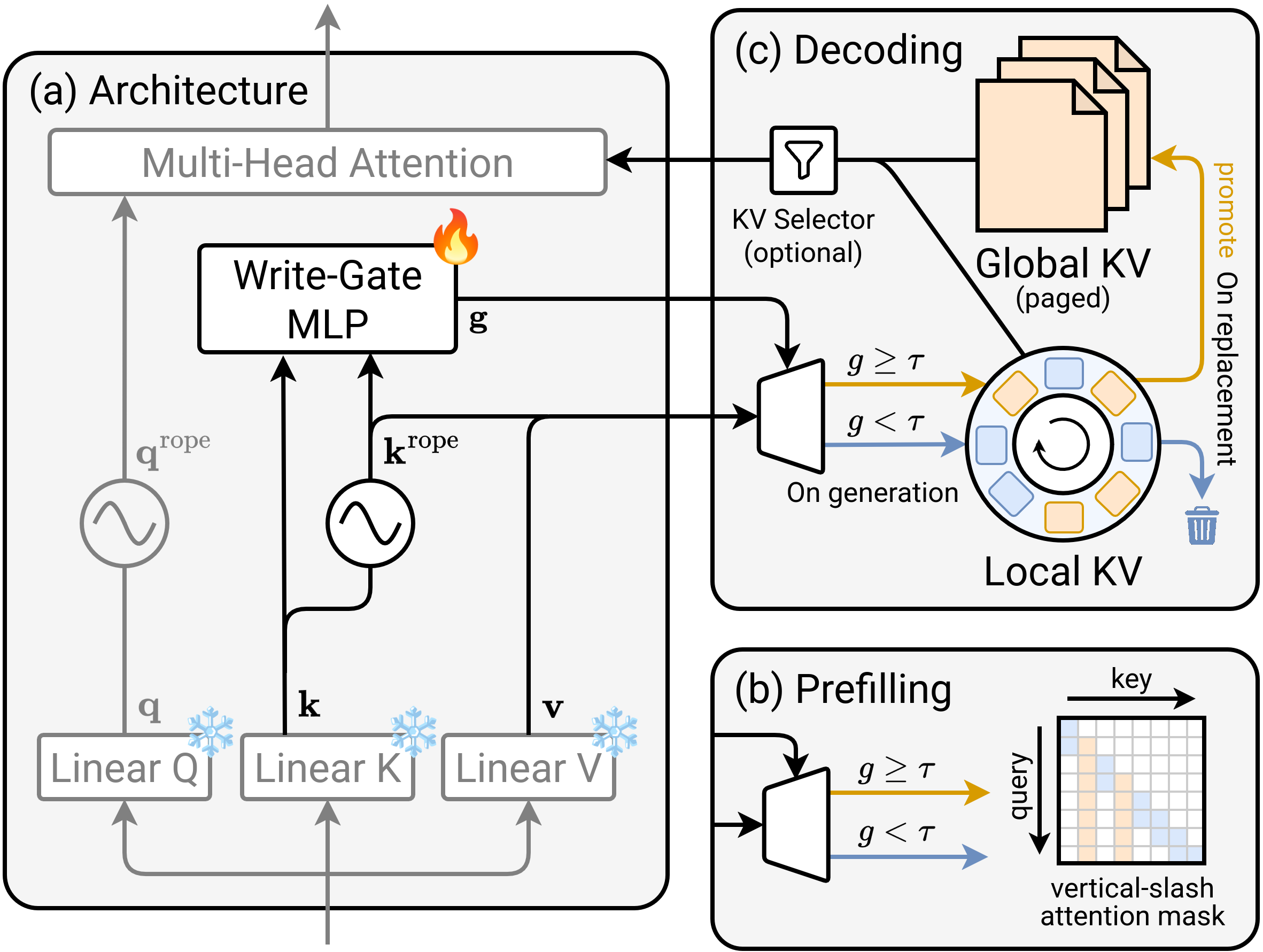}{Overview of Write-Gated KV}{}

\declareWideFig{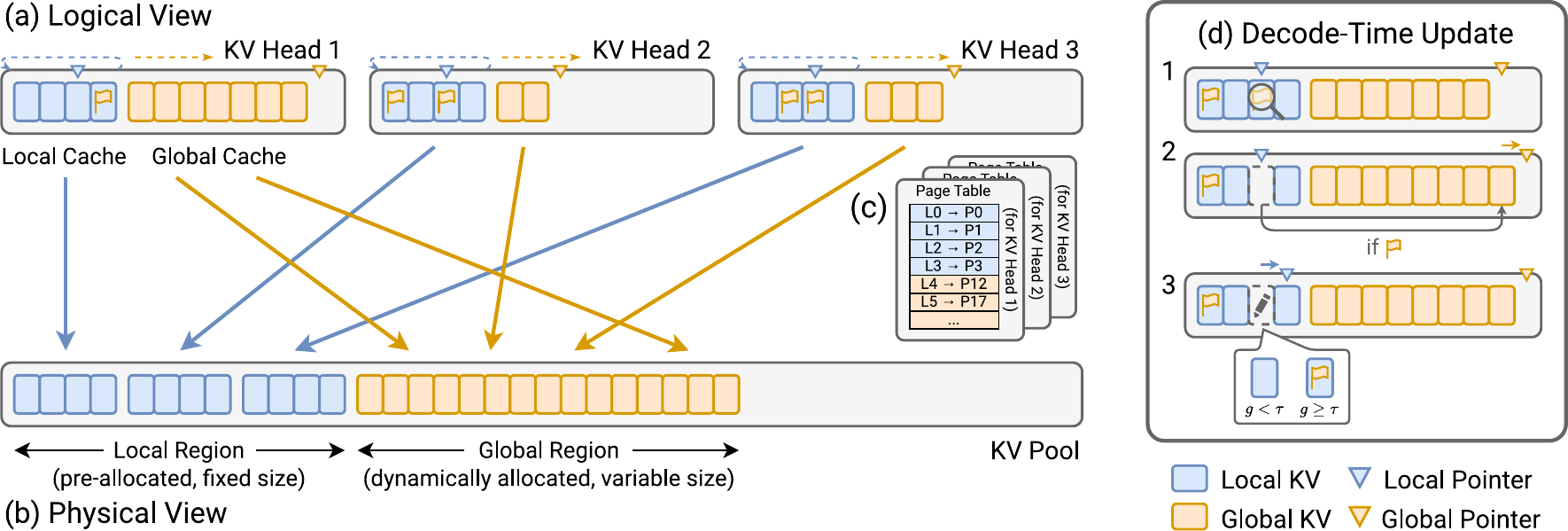}{Dual-cache paged memory management}{(a--c) To handle ragged cache growth across heads, we decouple the logical view into a fixed-size \textit{local cache} and a growing \textit{global cache}, mapped to a unified physical \textit{KV pool} via per-head \textit{page tables} to prevent memory fragmentation. (d) The cache update logic during decoding. We employ \textit{lazy promotion} where the token exiting the local window is inspected (and possibly promoted) upon replacement.}

\declareWideFig{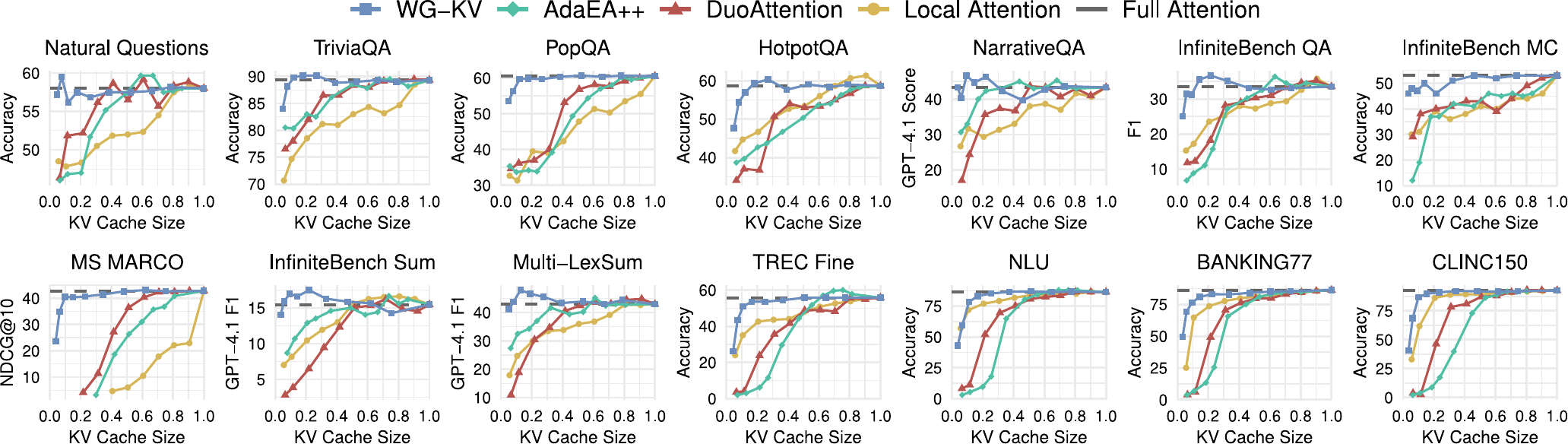}{Long-context HELMET task accuracy on Llama-3.1-8B}{The x-axis represents the normalized KV cache size. WG-KV consistently outperforms baselines under tight admission budgets.}

\declareFig{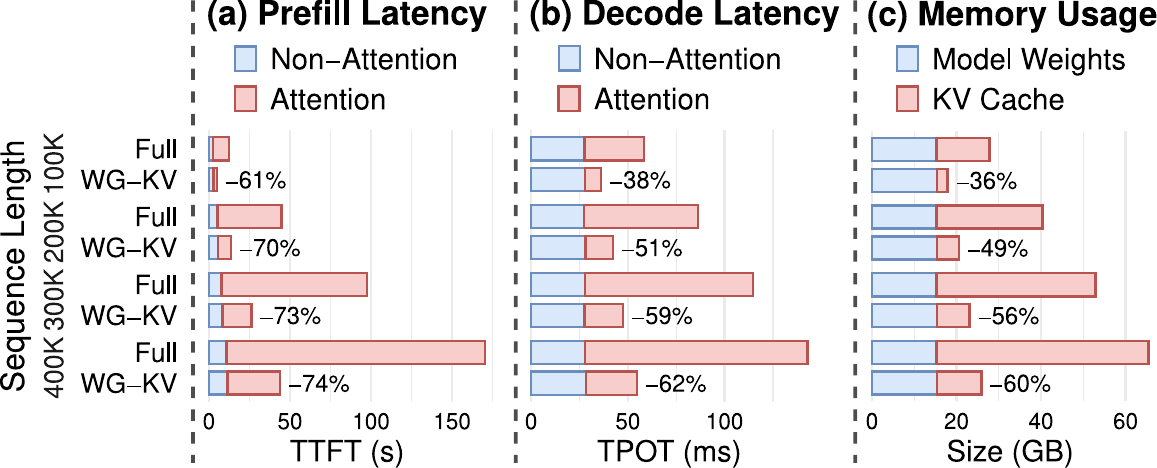}{End-to-end latency and memory usage on Llama-3.1-8B}{}

\declareRoundedFig{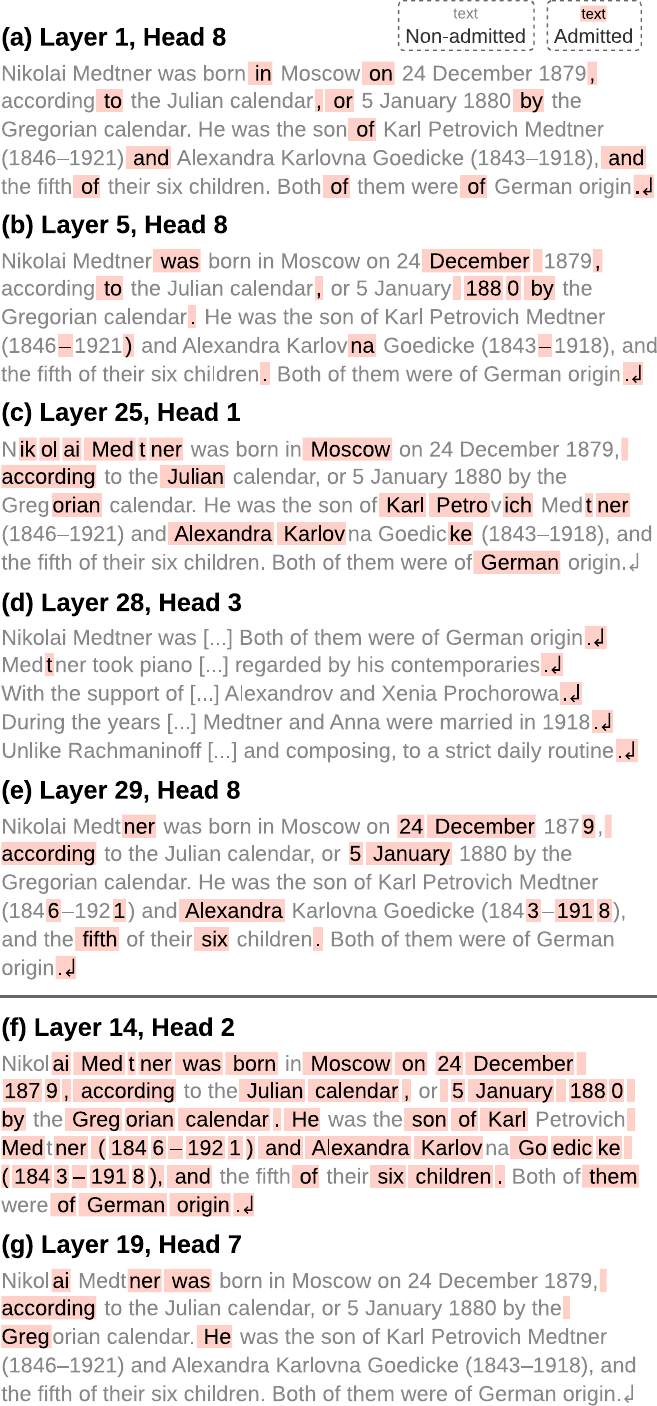}{5pt}{Visualization of admission decisions}{Highlighted tokens are admitted to the global cache.}

\declareWideFig{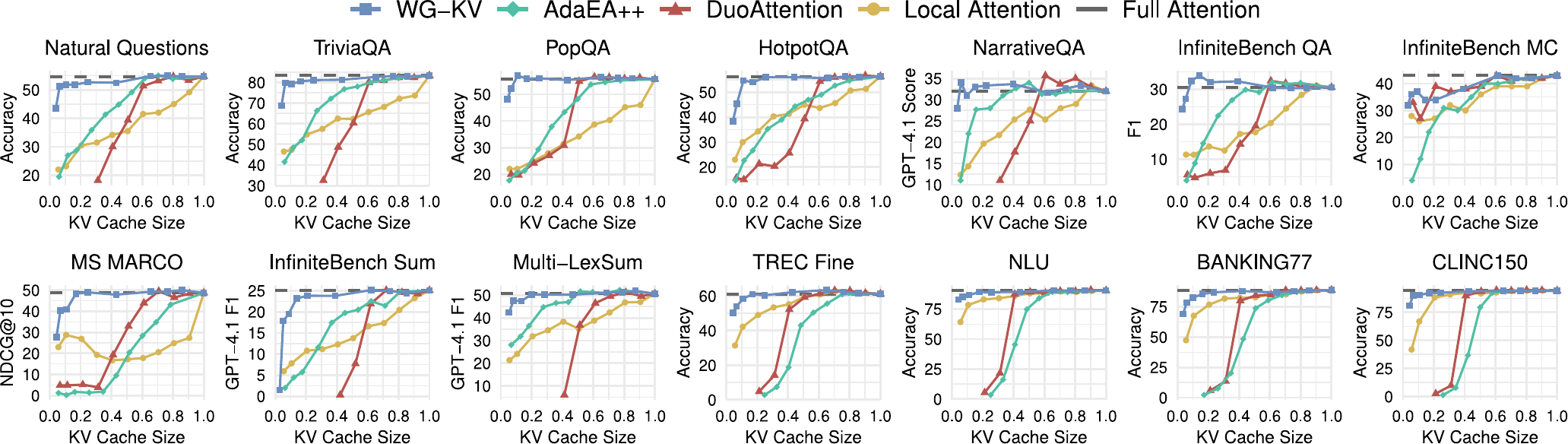}{Long-context HELMET task accuracy on Qwen3-4B-2507}{The trends align with the results on Llama-3.1-8B, showing WG-KV's generalizability across model architectures.}

\declareFig{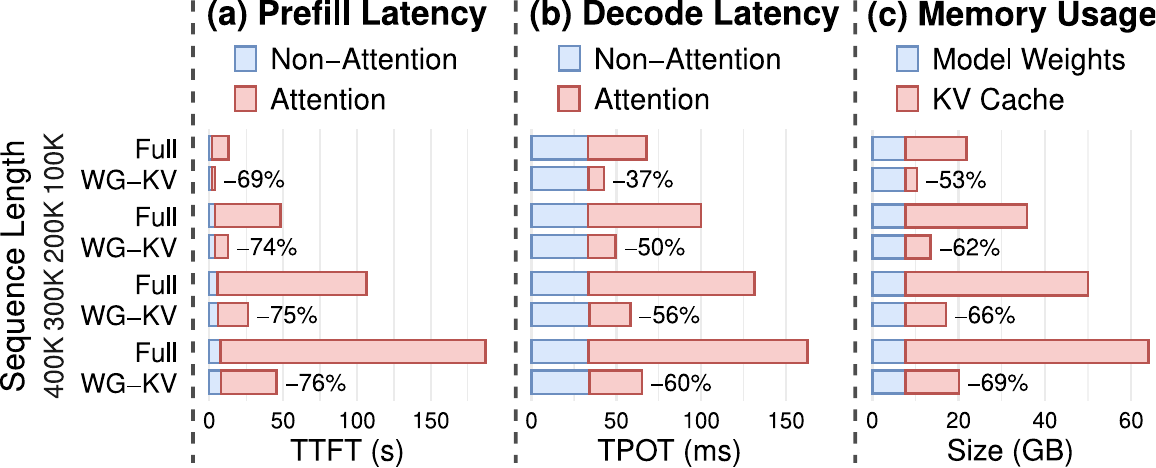}{End-to-end latency and memory usage on Qwen3-4B-2507}{}

\declareWideFig{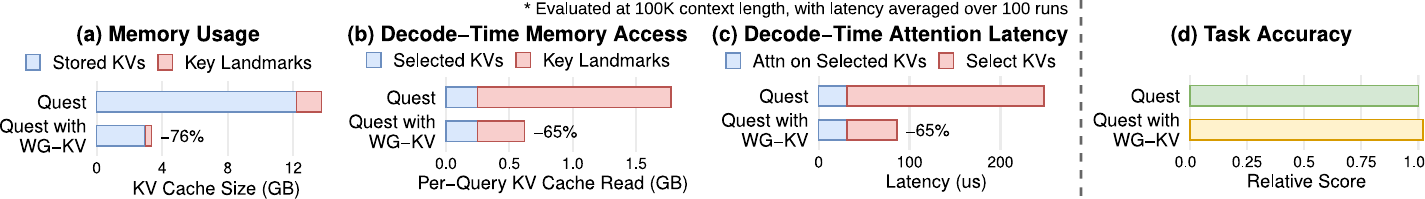}{Synergy between WG-KV and Quest}{By pre-filtering redundant tokens, WG-KV effectively (a) reduces KV memory usage. This compacted cache shrinks the candidate pool for selection, leading to (b) lower decode-time memory access and (c) reduced attention latency. Furthermore, integrating WG-KV (d) maintains nearly identical task accuracy compared to standalone Quest.}

\declareWideFig{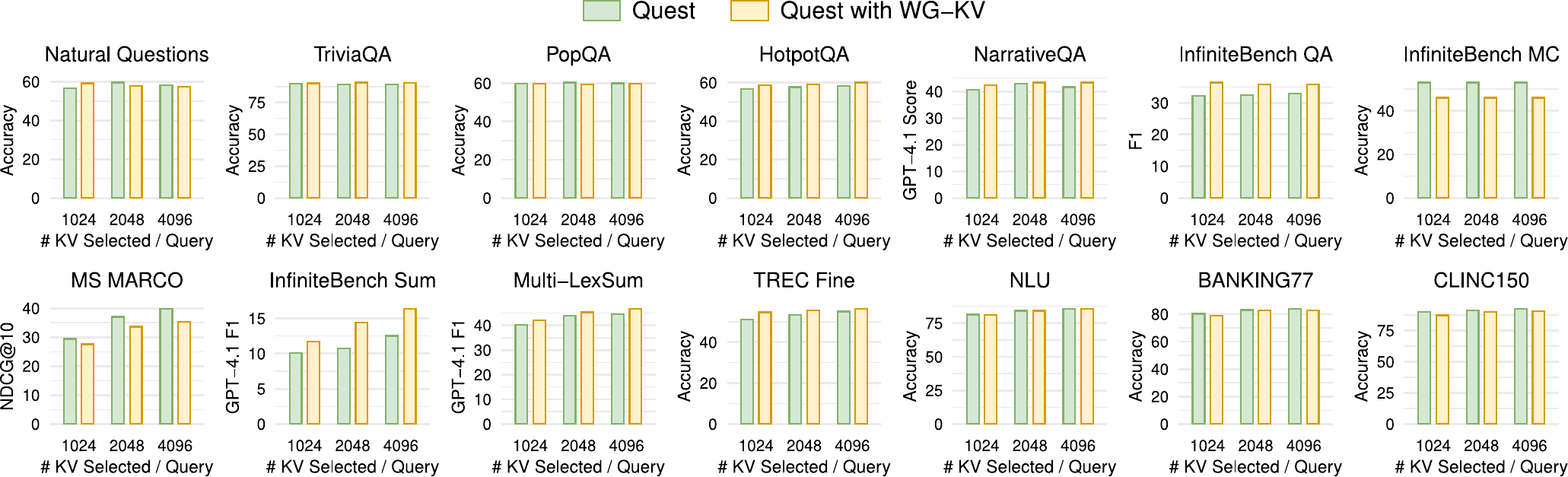}{Long-context HELMET task accuracy under varying selection budgets}{The results demonstrate that integrating WG-KV with Quest consistently matches the accuracy of standalone Quest, highlighting the composable nature of pre-write Admission and read-time Selection.}

\declareFig{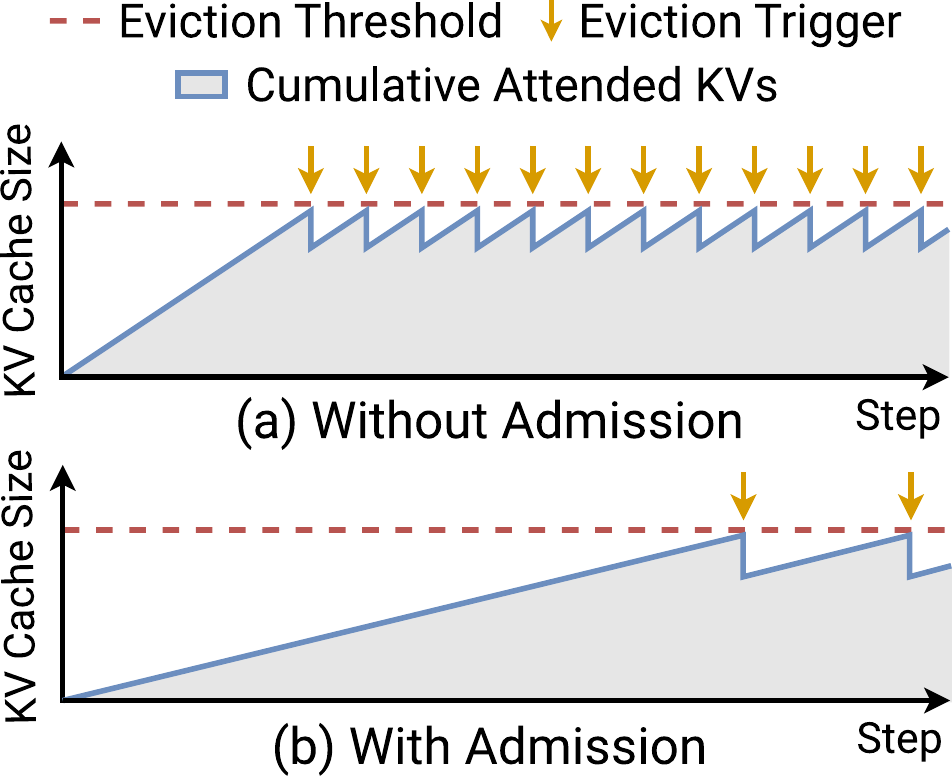}{Conceptual illustration of the synergy between KV Admission and Eviction}{}

\declareFig{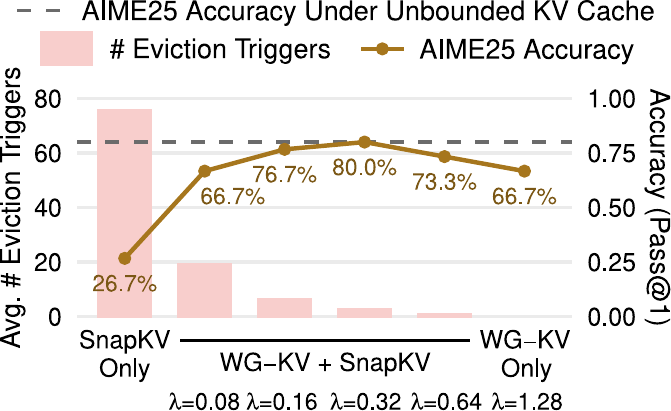}{Synergy between WG-KV and SnapKV on AIME25 under strict memory bounds}{}

\declareFig{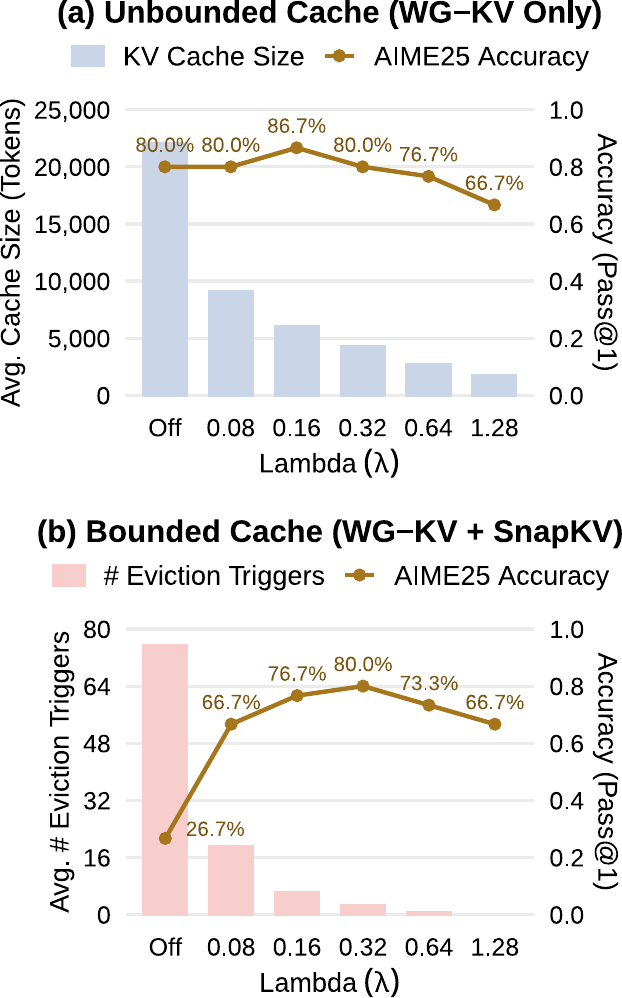}{Admission-Eviction synergy on AIME25 using Qwen3-4B-2507}{``Off'' denotes the baseline where WG-KV is disabled, and the reported metrics (KV Cache Size and \# Eviction Triggers) are averaged across all samples.}

\declareRoundedFig{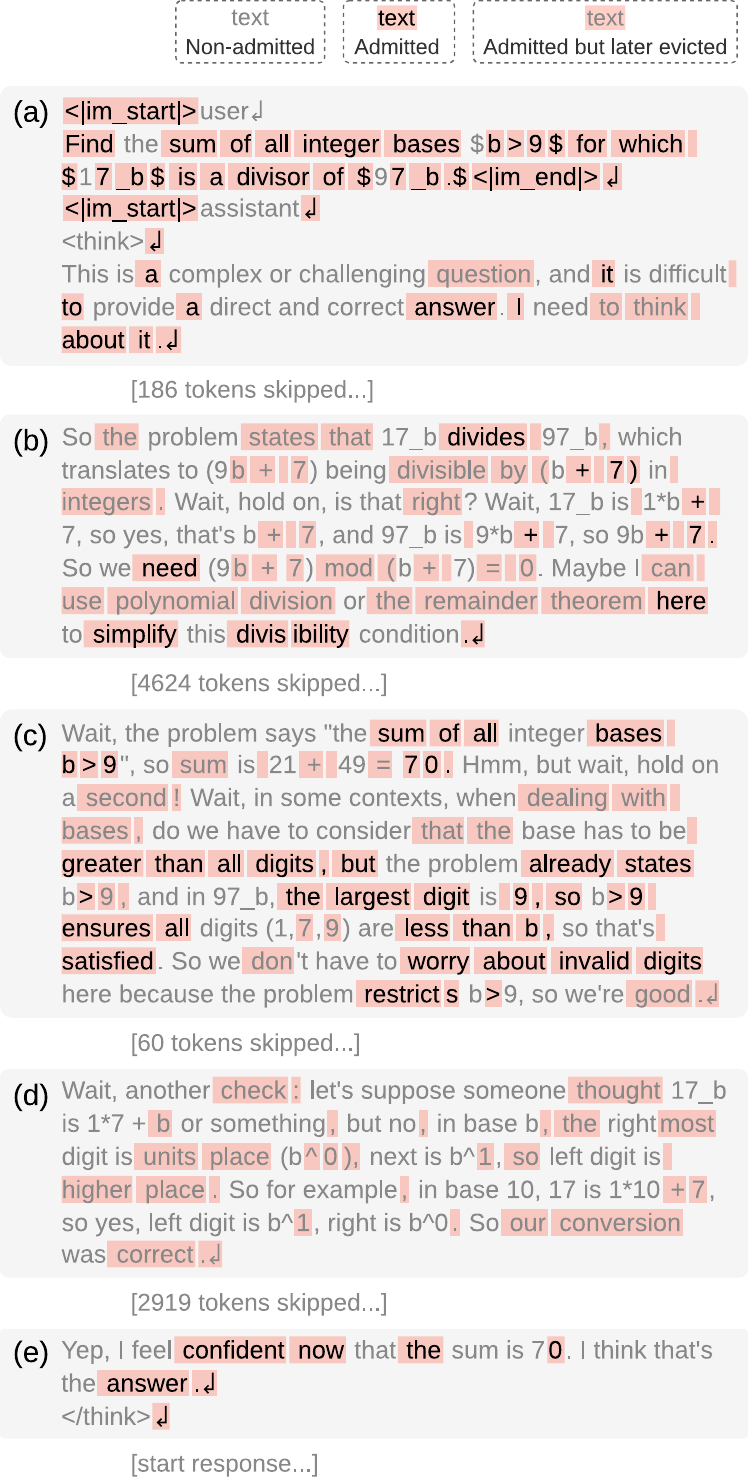}{5pt}{Visualization of the token lifecycle during an AIME25 reasoning trace using Qwen3-4B-2507 on Layer 25, Head 3}{}

\declareFig{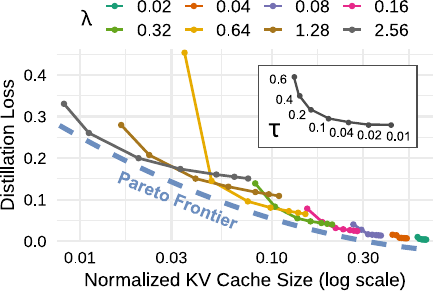}{Impact of $\lambda$ and $\tau$ on sparsity-loss tradeoff}{}

\declareFig{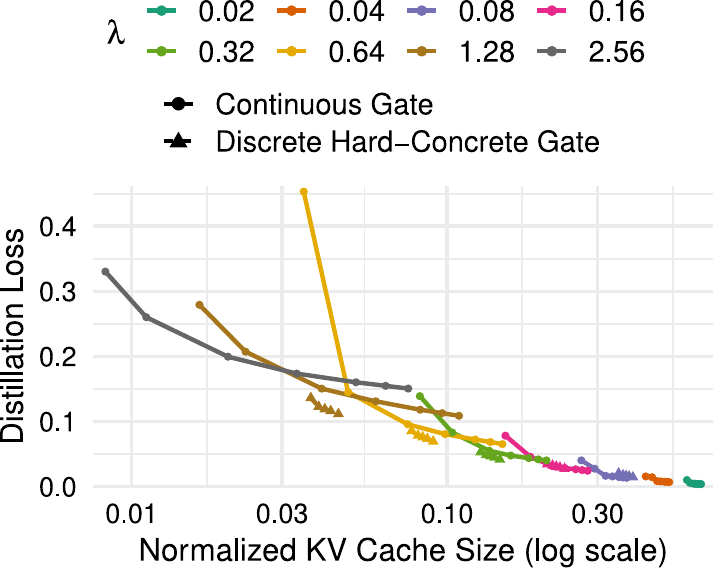}{Comparison of continuous and discrete gating on the sparsity-loss tradeoff}{}

\declareFig{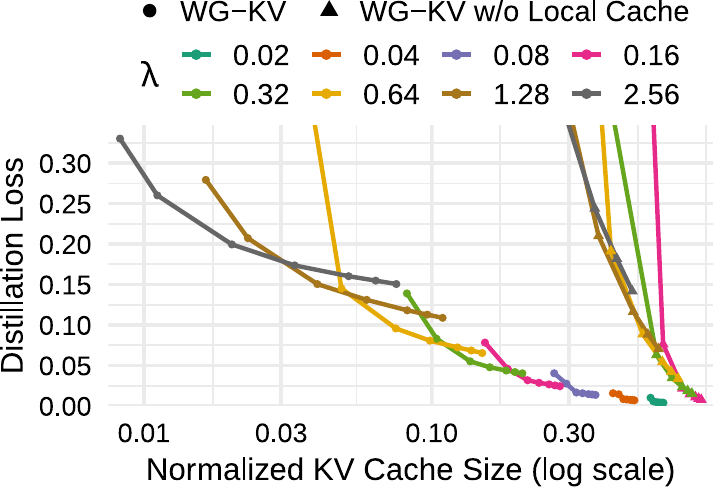}{Impact of local cache on the sparsity-loss tradeoff}{Removing local cache leads to a sharp surge in distillation loss.}

\declareWideFig{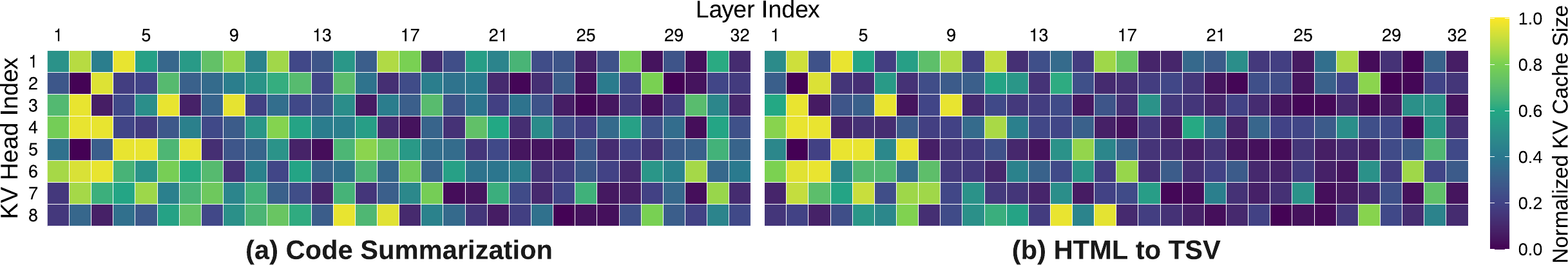}{Visualization of input-dependent admission patterns}{The heatmaps show the normalized KV cache size across each head for Llama-3.1-8B with WG-KV. We compare the \textit{(a) Code Summarization} task, which exhibits a relatively uniform distribution, against the \textit{(b) HTML to TSV} task, which shows a sparser pattern. This contrast demonstrates that WG-KV learns an input-dependent policy based on the semantic structure of the task.}

\declareRoundedWideFig{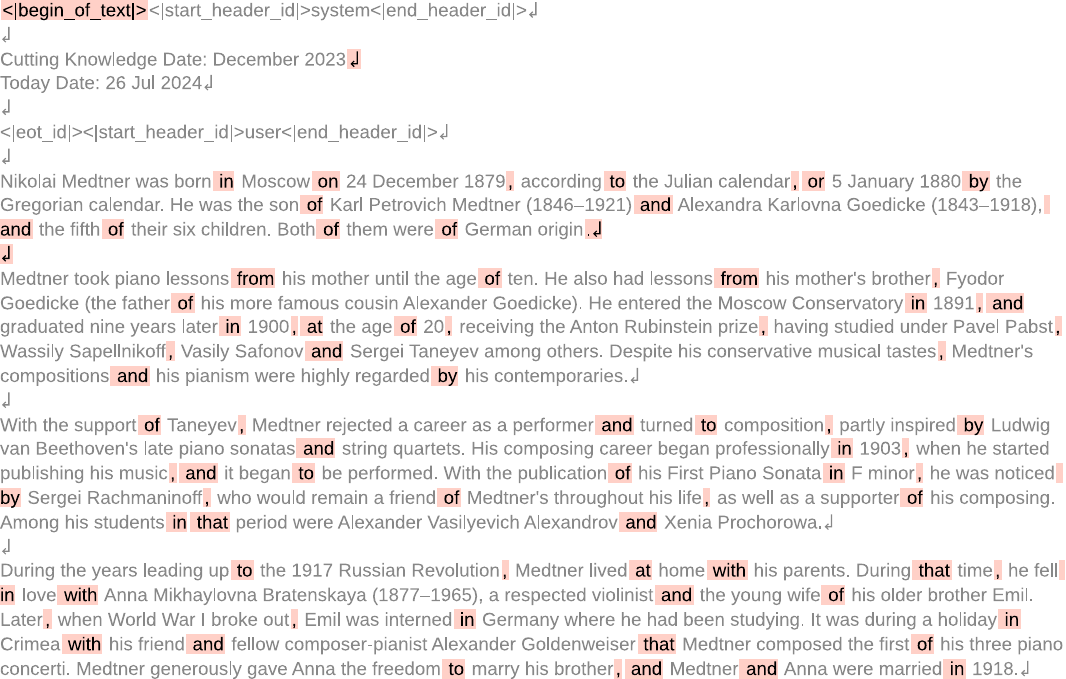}{10pt}{Full visualization of admission decisions for Layer 1, Head 8}{}

\declareRoundedWideFig{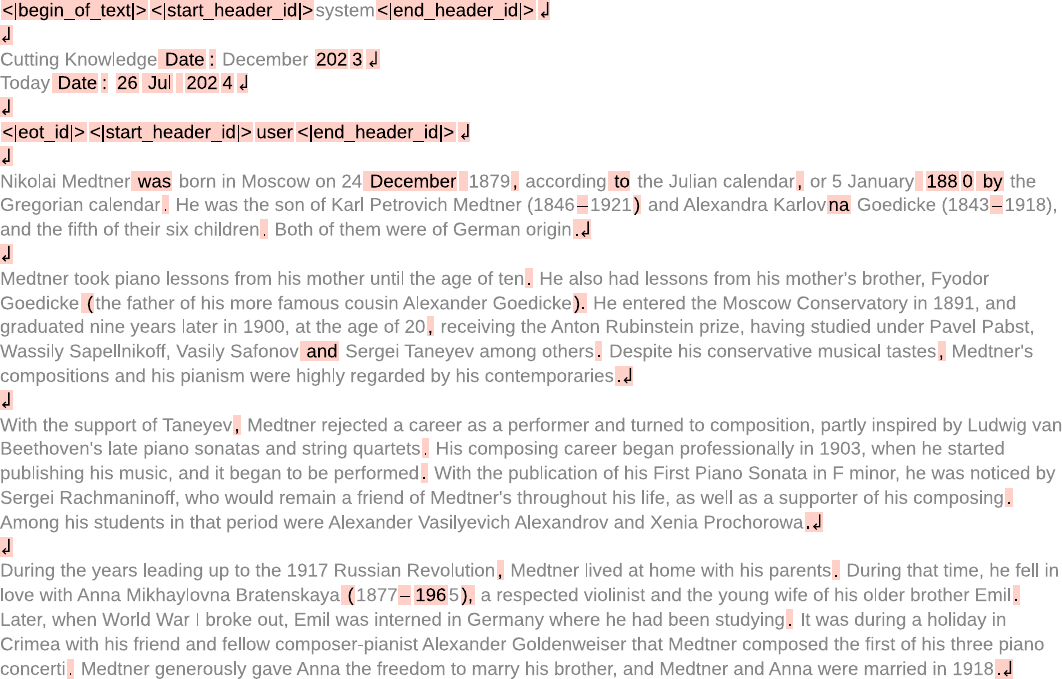}{10pt}{Full visualization of admission decisions for Layer 5, Head 8}{}

\declareRoundedWideFig{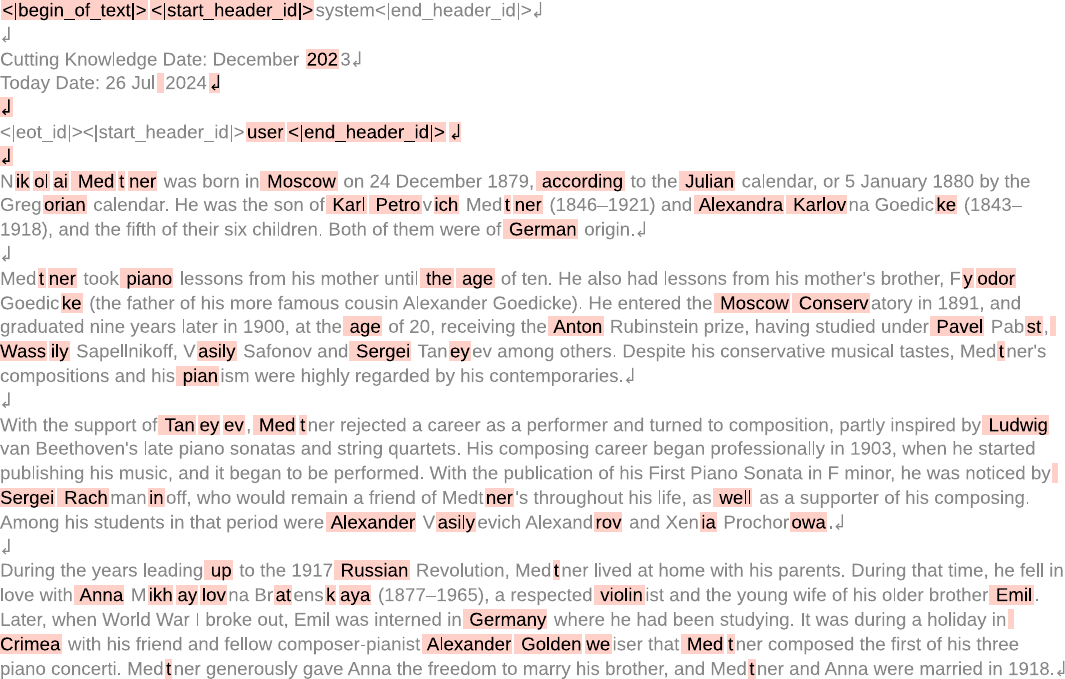}{10pt}{Full visualization of admission decisions for Layer 25, Head 1}{}

\declareRoundedWideFig{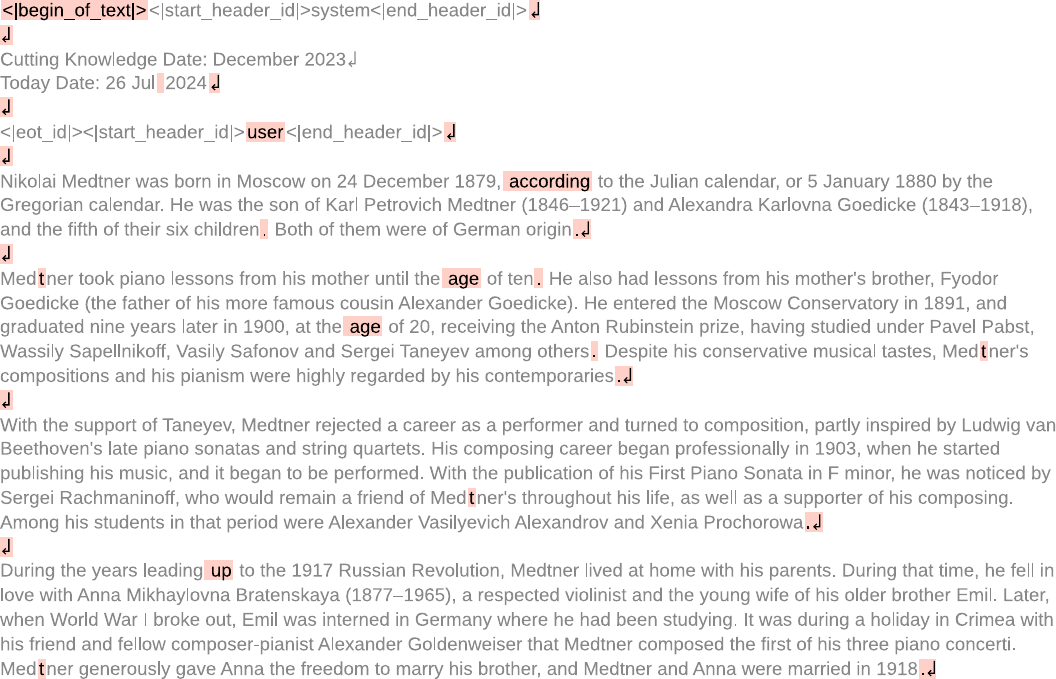}{10pt}{Full visualization of admission decisions for Layer 28, Head 3}{}

\declareRoundedWideFig{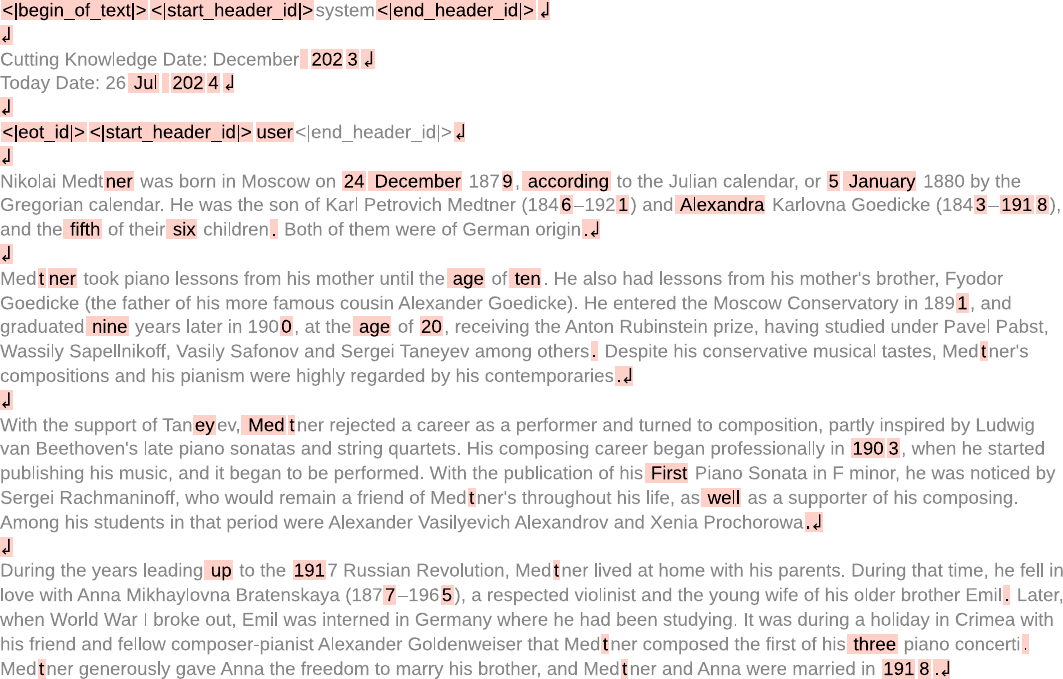}{10pt}{Full visualization of admission decisions for Layer 29, Head 8}{}

\declareRoundedWideFig{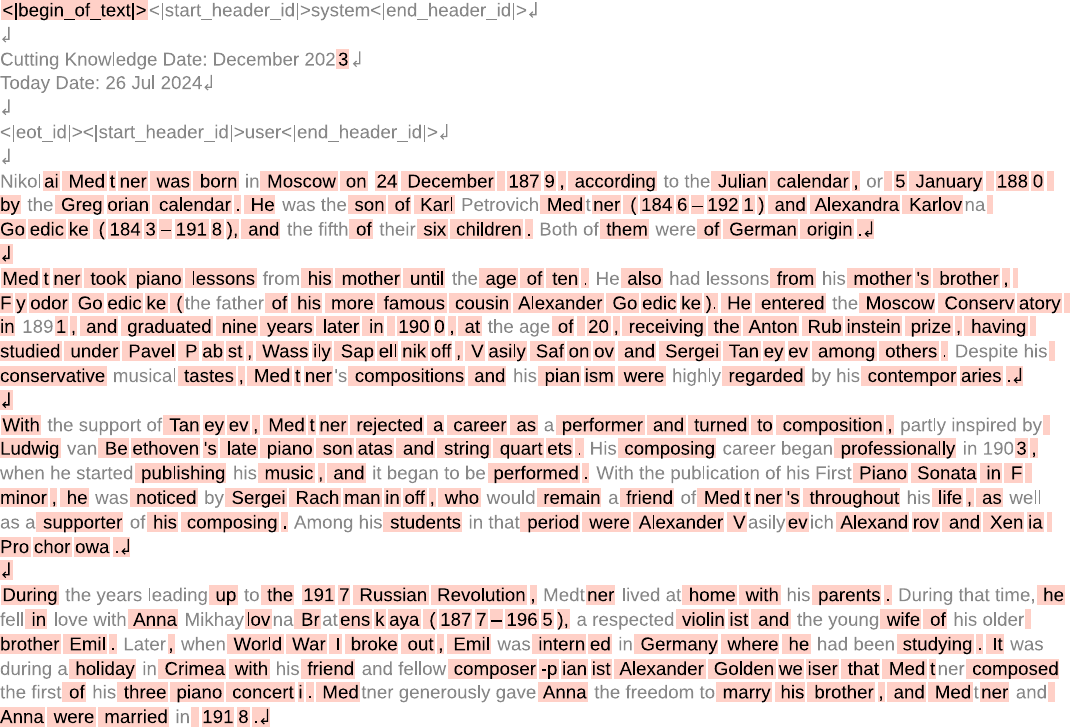}{10pt}{Full visualization of admission decisions for Layer 14, Head 2}{}

\declareRoundedWideFig{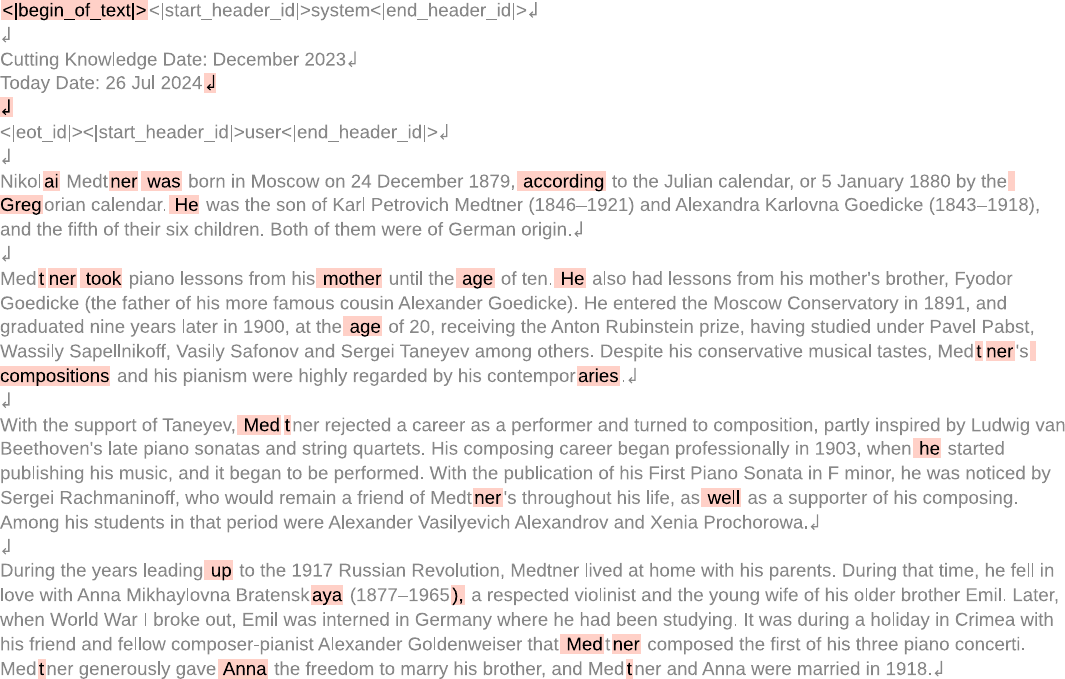}{10pt}{Full visualization of admission decisions for Layer 19, Head 7}{}

\declareRoundedWideFig{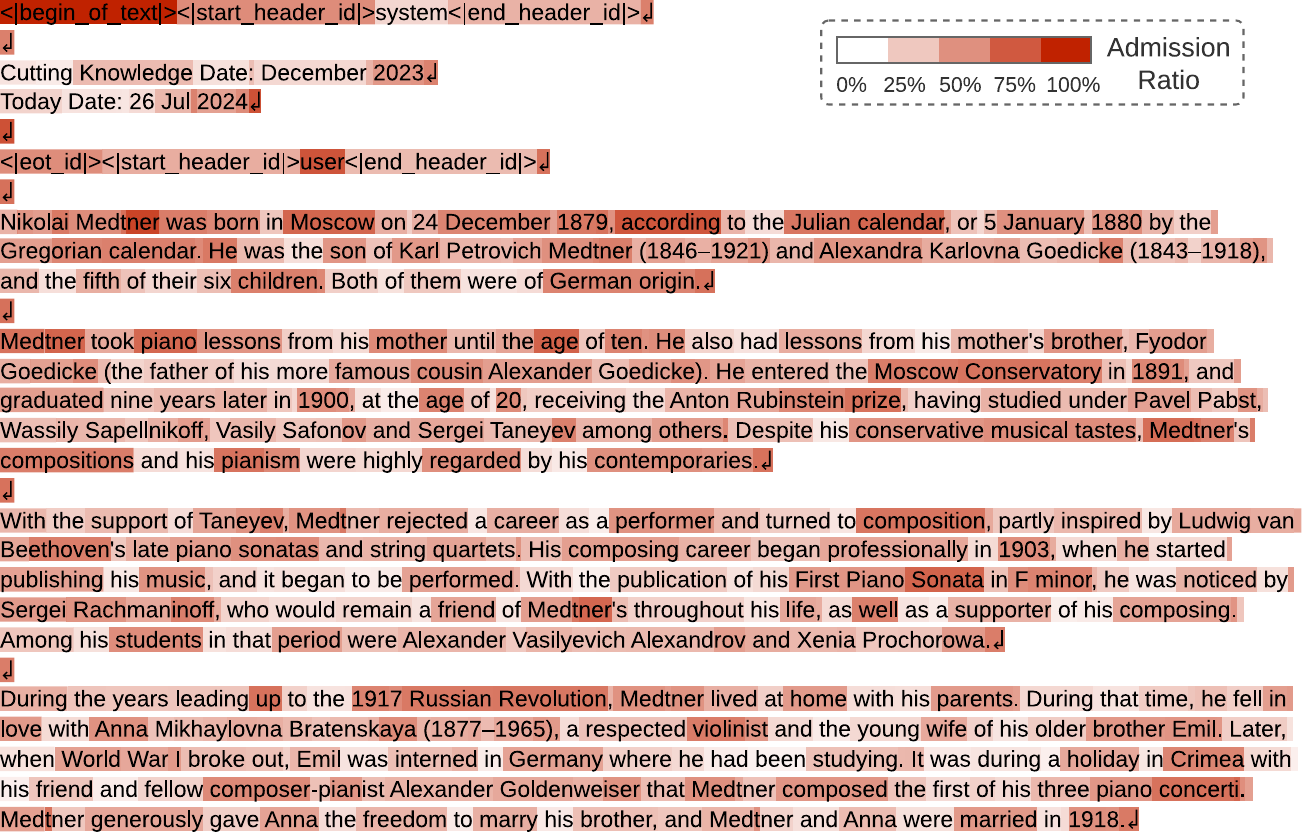}{10pt}{Aggregated admission decisions across all attention heads}{Color intensity denotes the proportion of heads admitting each token. High cross-head consensus is observed on structural control tokens, key entities, and semantically dense words, whereas function words and pronouns exhibit lower admission ratios.}

\declareTab{replay}
  {Average last-layer hidden states MSE under different admission policies}
  {}
  {
    \begin{tabular}{ccc}
      \toprule
      \textbf{Own Policy} & \textbf{Replay Policy} & \textbf{Random Policy} \\ 
      \midrule
      0.01480 & 0.10297 & 0.10271 \\ 
      \bottomrule
    \end{tabular}
  }
\setlist[itemize]{leftmargin=1.15em}
\setlist[enumerate]{leftmargin=1.15em}

\title{KV Admission: Learning What to Write for Efficient Long-Context\\LLM Inference}

\author{
 Yen-Chieh Huang\textsuperscript{1,2},
 Pi-Cheng Hsiu\textsuperscript{2,$\dagger$},
 Rui Fang\textsuperscript{1},
 Ming-Syan Chen\textsuperscript{1,$\dagger$}
\\
\\
 \textsuperscript{1}Department of Electrical Engineering, National Taiwan University
\\
 \textsuperscript{2}Research Center for Information Technology Innovation, Academia Sinica
\\
 \small{
   \textsuperscript{$\dagger$}Correspondence: \href{mailto:pchsiu@citi.sinica.edu.tw}{pchsiu@citi.sinica.edu.tw}, \href{mailto:mschen@ntu.edu.tw}{mschen@ntu.edu.tw}
 }
}

\begin{document}
\maketitle

\begin{abstract}
Long-context LLM inference is bottlenecked by the quadratic attention complexity and linear Key-Value (KV) cache growth. Prior approaches mitigate this via post-hoc selection or eviction but overlook the root inefficiency: \textit{indiscriminate token admission}. In this paper, we formalize KV management as a causal system of three primitives: KV Admission, Selection, and Eviction. We instantiate KV Admission via Write-Gated KV (WG-KV), a lightweight mechanism that learns to predict token utility \textit{before} cache entry. By filtering out redundant states early to maintain a compact global cache alongside a sliding local cache, WG-KV significantly reduces memory usage and accelerates both prefill and decode phases. Our results demonstrate that \textit{learning what to write} is a principled and practical recipe for efficient long-context inference. Code is available at \url{https://github.com/EMCLab-Sinica/WG-KV}.
\end{abstract}

\ifabstractonly\else
\section{Introduction}

\myFloat{primitives}{t}{0.9}

The capability of Large Language Models (LLMs) to process extensive contexts has revolutionized applications ranging from long-document summarization \cite{longbenchv2}, repository-level code analysis \cite{coderepoqa}, to long-term agentic planning \cite{planandact}. However, this capability comes at a steep serving cost. Long-context LLM inference is bottlenecked by both the quadratic attention complexity and the linear growth of the Key-Value (KV) cache. In long-context scenarios, the sheer volume of tokens not only incurs massive computational overhead but also exhausts limited GPU memory. Consequently, the latency and resource consumption increase significantly with the context length, rendering long-context LLM inference prohibitively expensive.

To mitigate this bottleneck, prior research has focused on two management strategies: \textbf{KV Selection} and \textbf{KV Eviction}. Selection-based methods, such as Quest \cite{quest} and NSA \cite{nsa}, selectively attend to relevant KVs at runtime yet retain the complete state in memory. Eviction-based methods, such as SnapKV \cite{snapkv} and R-KV \cite{rkv}, retrospectively prune the cache by evicting unneeded tokens after they have accumulated in memory. While effective, these paradigms share a common, fundamental inefficiency: \textit{every generated token is indiscriminately admitted to the growing cache state}, where the system blindly consumes memory resources to accumulate redundant KV states, only to later skip or evict them.

In this paper, we argue that an efficient long-context system requires a third, missing primitive: \textbf{KV Admission}. Instead of retrieving information at \textit{read-time} (Selection) or managing memory \textit{post-write} (Eviction), Admission operates \textit{pre-write}, acting as a gatekeeper that predicts the future utility of a token before admitting it to the KV cache.

To realize this, we introduce \textbf{Write-Gated KV (WG-KV)}, a novel, learnable KV Admission mechanism. Unlike heuristic approaches, WG-KV employs a lightweight, end-to-end differentiable predictor that evaluates token utility \textit{before} cache entry. During prefilling, this predictor constructs a vertical-slash attention mask, ensuring high-utility tokens are visible to the entire sequence while others are accessible only locally. Simultaneously, it routes generated KV pairs to distinct memory regions: tokens predicted as high-utility are admitted to a persistent global cache, while others are confined to a transient local cache. This selective strategy seamlessly extends to the decoding phase, where we maintain this dual-cache structure to filter redundant states upon KV generation.

Our approach yields substantial efficiency gains through a \textbf{hardware-aware implementation} that effectively translates theoretical sparsity into practical wall-clock speedups. We evaluate WG-KV on Llama 3 \cite{llama3} and Qwen3 \cite{qwen3} models, demonstrating that WG-KV reduces memory usage by 36--69\% and delivers 2.56--4.17x prefill and 1.59--2.63x decode speedups, while maintaining near-lossless task accuracy across diverse long-context benchmarks. By \textit{learning what to write}, we enable LLMs to sustain longer contexts with significantly fewer resources.
\section{Background and Motivation}

\subsection{Long-context attention bottleneck}

The core of the Transformer architecture \cite{attn} in LLMs is the attention mechanism. Given a sequence of input tokens, the attention output for a query vector $q_t \in \mathbb{R}^d$ at step $t$ is computed by attending to past key-value pairs $\{(k_i, v_i)\}_{i=1}^{t}$. Specifically, let $K_{\leq t},V_{\leq t} \in \mathbb{R}^{t \times d}$ denote the matrices formed by stacking the key and value vectors up to step $t$. The attention operation is defined as:
\begin{myEquation}
    \operatorname{Attn}(q_t, K_{\leq t}, V_{\leq t}) = \operatorname{Softmax}\left(q_t K_{\leq t}^\top/\sqrt{d}\right) V_{\leq t}
\end{myEquation}

In standard autoregressive generation, recomputing $K_{\leq t}$ and $V_{\leq t}$ from all previous tokens at every step $t$ would be computationally prohibitive. To address this, inference engines employ a \textbf{KV cache}, storing the key and value vectors for previous tokens. At each step $t$, the system only computes the new pair $(k_t, v_t)$, appends it to the cache, and performs attention over the accumulated state. However, as the sequence length increases, the attention mechanism presents distinct bottlenecks during the two phases of inference: \textit{prefill} and \textit{decode}.

\textit{During prefilling}, the model processes the sequence of length $N$ in parallel, with the quadratic complexity of attention $O(N^2)$ quickly dominates execution time (\cref{fig:breakdown}a). \textit{During decoding}, the model generates tokens autoregressively, making the process memory-bound as each step requires loading the KV cache from memory. As the sequence grows, the massive data transfer saturates memory bandwidth, driving up latency (\cref{fig:breakdown}b), while the expanding cache occupies significant GPU memory (\cref{fig:breakdown}c).

\myFloat{breakdown}{t}{1}

\subsection{A causal framework for KV management}
\label{subsec:framework}

Formally, we can model the KV cache $\mathcal{C}_t$ for a certain attention head at step $t$ as a set of KV pairs stored in memory. In standard LLM inference, the cache management policy is trivial: it is an append-only operation where every new token's state $(k_t, v_t)$ is indiscriminately accumulated throughout the entire inference process:
\begin{myEquation}
    \mathcal{C}_t^{\text{std}} = \mathcal{C}_{t-1}^{\text{std}} \cup \{(k_t, v_t)\}
\end{myEquation}

This policy leads to linear KV cache growth $|\mathcal{C}_t^{\text{std}}| = t$ and becomes the root cause of attention bottleneck. To address this, prior works have introduced constraints on the attention operation or $\mathcal{C}_t$ itself. We unify these approaches under a causal framework based on the \textit{timing} and \textit{decision scope} of the operation, as illustrated in \cref{fig:primitives} and detailed below.

\paragraph{KV Selection (read-time).}
Selection methods, such as Quest \cite{quest} and NSA \cite{nsa}, optimize the \textit{read} path. They \textit{approximate} the attention operation by selecting a subset of keys $\mathcal{S}_t \subset \mathcal{C}_t$ relevant to the current query $q_t$:
\begin{myEquation}
  \operatorname{Attn}(q_t, \mathcal{C}_t) \approx \operatorname{Attn}(q_t, \mathcal{S}_t),\; \text{where } \mathcal{S}_t = \operatorname{Select}(q_t, \mathcal{C}_t)
\end{myEquation}

While Selection reduces the number of attended tokens per step, the underlying cache $\mathcal{C}_t$ remains fully populated to support future queries, leaving the linear KV cache growth unresolved.

\paragraph{KV Eviction (post-write).}
Eviction methods, such as SnapKV \cite{snapkv} and R-KV \cite{rkv}, impose a constraint $B$ on the cache size. These methods operate \textit{retrospectively}: they first admit the new token into the cache, and then prune the set based on \textit{past statistics}:
\begin{myEquation}
    \mathcal{C}_t^{\text{new}} = \operatorname{Evict}(\mathcal{C}_t^{\text{old}}) \quad \text{s.t.} \quad |\mathcal{C}_t^{\text{new}}| \leq B
\end{myEquation}

Eviction effectively bounds KV memory usage. However, it suffers from a ``write-then-throw'' inefficiency: the system temporarily admits noise into the KV cache, forcing subsequent queries to waste memory bandwidth and computation attending to these tokens until they are eventually identified as redundant. Furthermore, eviction can be risky; information discarded is permanently lost, potentially degrading generation quality if the evicted tokens are needed for later context \cite{infinigen}.

\paragraph{KV Admission (pre-write).}
We identify \textit{Admission} as the missing third primitive. Unlike Eviction, which reacts to cache overflow, Admission acts as a gatekeeper. It evaluates the \textit{future utility} of a new state $(k_t, v_t)$ \textit{before} it is committed to $\mathcal{C}_t$:
\begin{myEquation}
    \mathcal{C}_t = \operatorname{Admit}(\mathcal{C}_{t-1}, k_t, v_t)
\end{myEquation}

Ideally, an admission policy prevents the persistence of noise tokens, ensuring $\mathcal{C}_t$ holds only critical information. This avoids the cost of accumulating redundant states, reducing KV growth and alleviating subsequent attention bottlenecks.

\paragraph{A unified view.}

As illustrated in \cref{fig:primitives}, these primitives are not mutually exclusive; rather, they operate at different stages of the token lifecycle. An ideal inference system can employ Admission to \textit{filter} the input stream, Selection to \textit{focus} the attention computation, and Eviction to \textit{prune} obsolete history. Together, they form a comprehensive framework that addresses the inefficiencies of long-context LLM inference across the entire token lifecycle. We provide a more comprehensive survey of existing literature in Appendix~\ref{app:related}.

\subsection{Sparsity and locality in attention patterns}
\label{subsec:sparsity}

To design an effective KV Admission policy, we must first understand the characteristics of attention patterns. We employ Llama-3.1-8B \cite{llama3} to perform a long-context code summarization task from The Stack dataset \cite{thestack} to observe \textit{which} tokens are attended to and \textit{when}. We identify three key properties:

\myFloat{distribution}{t}{1}

\paragraph{Skewed utility distribution.}
Attention scores are not uniformly distributed; they exhibit pronounced sparsity where a small subset of tokens attracts the majority of attention. As shown in \cref{fig:distribution}a, specific tokens (e.g., Token 383 in Layer 10, Head 19) act as high-utility tokens, consistently receiving high attention from subsequent queries. In contrast, other tokens (e.g., Token 1185 in the same head) are virtually ignored. This suggests that a large portion of the KV cache stores ``noise'' that contributes negligible value to future representations.

\paragraph{Head-specific relevance.}
The utility of a token is not intrinsic to the token itself but is dependent on the attention head. A token that is critical for one head may be irrelevant for another. For instance, comparing the two heads in \cref{fig:distribution}b, we observe that while Token 1185 serves as a primary attention target for Layer 22, Head 6, it is almost completely ignored by Layer 10, Head 19. Conversely, Token 383, which was ignored by the former, becomes highly salient for the latter. This implies that a unified admission policy (e.g., dropping a token from all heads simultaneously) is suboptimal. An effective admission policy must be fine-grained, making independent decisions for each head.

\paragraph{The necessity of local context.} While long-range attention is sparse, local attention exhibits strong recency bias. As illustrated in \cref{fig:distribution}c, Token 383 in Layer 22, Head 6 exemplifies a phenomenon of ``transient utility.'' Although this token eventually receives near-zero attention from distant queries, it attracts significant attention from its immediate successors. This creates a dilemma for any admission policy: a token cannot be immediately dismissed as universally irrelevant, as it exhibits strong local saliency. A hybrid mechanism is therefore needed to preserve recent tokens briefly for local dense attention before switching to utility-based admission for long-range context.

\subsection{Systems challenges with ragged KV states}
\label{subsec:ragged}

\myFloat{fragmentation}{b}{0.93}

Implementing a fine-grained admission policy introduces non-trivial systems challenges. In standard dense attention, the KV cache for a sequence is represented as a contiguous tensor. However, independent admission decisions by each head, as motivated in \cref{subsec:sparsity}, result in a cache that is ``ragged'' across two dimensions: \textit{(1) Across heads:} Different heads prioritize different tokens, leading to uneven cache lengths within the same layer. \textit{(2) Across layers:} The level of semantic abstraction varies by depth, causing sparsity rates to diverge between shallow and deep layers.

A naive implementation that simply concentrates admitted tokens leads to severe memory fragmentation, as shown in \cref{fig:fragmentation}. Pre-allocating maximum-length buffers negates the memory savings, while dynamic reallocation incurs prohibitive runtime overhead. To translate theoretical sparsity into practical wall-clock speedup and memory reduction, the system must leverage a decoupled logical-physical memory mapping to manage these head-wise irregular caches efficiently.
\section{Method}
In this section, we present \textbf{Write-Gated KV (WG-KV)}, a learnable KV Admission mechanism designed to address the aforementioned inference inefficiencies. We first introduce the architectural overview of WG-KV (\cref{subsec:arch}). Then, we derive the differentiable gating mechanism that enables end-to-end learning of token utility (\cref{subsec:gate}). Finally, we present the training objective for learning the admission policy (\cref{subsec:objective}).

\subsection{Architectural overview}
\label{subsec:arch}

\myFloat{overview}{t}{1}

The core philosophy of WG-KV is to filter tokens at the source by \textit{determining which KV states merit long-term retention}. We introduce a lightweight \textbf{Write-Gate MLP} (\cref{fig:overview}a) into the standard attention block to evaluate the gating score $g_{t} \in [0, 1]$ for each token. Based on this score, the system selectively routes the KV pair to one of two logical memory regions: \textit{(1) Global cache:} A memory region for tokens with high gating score ($g_{t} \approx 1$). These tokens are retained persistently to support long-term dependencies. \textit{(2) Local cache:} A sliding window of size $W_{\text{local}}$ that unconditionally retains recent tokens to support their transient utility (\cref{subsec:sparsity}). Upon exiting this window, tokens with low gating score ($g_{t} \approx 0$) are discarded.

\subsection{Differentiable gating mechanism}
\label{subsec:gate}

To learn the gating score $g$, we derive it using the MLP and subsequently integrate it into the attention formulation. Specifically, for layer $l$, head $h$, token $t$, the input feature $x_{l,h,t}$ is defined by concatenating the pre-RoPE key $k_{l,h,t}$ and the post-RoPE key $k_{l,h,t}^{\text{rope}}$, with both terms normalized via RMSNorm \cite{rope,rmsnorm}:\footnote{In the context of Grouped-Query Attention \cite{gqa}, the term ``head'' specifically refers to the Key-Value (KV) head. We use ``head'' for simplicity throughout the paper.}
\begin{myEquation}
    x_{l,h,t} = \left[\operatorname{RMSNorm}(k_{l,h,t}); \operatorname{RMSNorm}(k_{l,h,t}^{\text{rope}})\right]
\end{myEquation}

The gating score $g_{l,h,t}$ is then formulated as:
\begin{myEquation}
    g_{l,h,t} = \sigma\left(W^2_{l,h} \cdot \operatorname{GELU}(W^1_{l,h} \cdot x_{l,h,t} + b^1_{l,h}) + b^2_{l,h}\right)
\end{myEquation}%
where $W^1_{l,h}, b^1_{l,h},W^2_{l,h}, b^2_{l,h}$ are learnable parameters, and $\sigma$ is the sigmoid function.

\paragraph{Write-Gated Attention.}
Standard Scaled Dot-Product Attention computes the output for a query $q_i \in \mathbb{R}^d$ against $\{(k_j,v_j)\}_{j \le i}$ as:\footnote{For notational simplicity, we omit the layer ($l$) and head ($h$) subscripts in variables like $q_t$, $k_t$, $v_t$ and $g_t$.}
\begin{myEquation}
    \operatorname{Attn}(q_i, K_{\leq i}, V_{\leq i}) = \sum_{j \le i} \frac{\exp(q_i \cdot k_j / \sqrt{d})}{\sum_{p \le i} \exp(q_i \cdot k_p / \sqrt{d})} \, v_j
\end{myEquation}

To simulate the effect of admission during training, we modulate attention scores via mask $m_{ij}$ that provides full visibility within the local window $W_{\text{local}}$ while weighting distant tokens by $g_j$:
\begin{myEquation}
m_{ij} = \max(\mathbb{1}(i - j < W_{\text{local}}), g_j)
\end{myEquation}%
where $\mathbb{1}$ is the indicator function. Our proposed Write-Gated Attention is then formulated as:
\begin{myEquation}
    \widetilde{\operatorname{Attn}}(q_i, K_{\leq i}, V_{\leq i}) = \sum_{j \le i} \frac{\exp(q_i \cdot k_j / \sqrt{d}) \cdot m_{ij}}{\sum_{p \le i} \exp(q_i \cdot k_p / \sqrt{d}) \cdot m_{ip}} \, v_j
\end{myEquation}

This formulation ensures that if $g_j=0$, the token $j$ effectively vanishes from the attention of distant queries, mimicking the behavior of not admitting it to the global cache. Conversely, if $g_j=1$, the token remains visible to all future queries, simulating its admission to the global cache.

\paragraph{Log-space transformation for efficient training.}
While the multiplicative gating above is theoretically sound, standard attention kernels \cite{flashattn} are optimized for $\exp(q \cdot k)$. Introducing a per-token multiplication factor disrupts standard kernel optimizations. To address this, we transform the gating operation into a log-space bias:
\begin{myEquation}
    \exp\left(q_i \cdot k_j / \sqrt{d}\right) \cdot m_{ij} = \exp\left(q_i \cdot k_j / \sqrt{d} + \log(m_{ij})\right)
\end{myEquation}

Write-Gated Attention is thus reformulated as:
\begin{myEquation}
    \widetilde{\operatorname{Attn}}(q_i, K_{\leq i}, V_{\leq i})=\operatorname{Softmax}\left( q_iK_{\leq i}^\top/\sqrt{d} + B_{\text{gate}} \right) V_{\leq i}
\end{myEquation}%
where $B_{\text{gate}}$ is an attention bias matrix derived from $\log(m_{ij})$. In practice, we compute $\log(m_{ij} + \epsilon)$ with small $\epsilon$ for numerical stability. This log-space transformation allows us to leverage generalized attention kernels, such as FlexAttention \cite{flexattn}, for efficient training on long sequences.

\subsection{Training objective}
\label{subsec:objective}

To learn the admission policy, we optimize a joint objective, with the total loss $\mathcal{L}_{\text{total}}$ defined as:
\begin{gather*}
\adjustbox{max width=\linewidth}{$\mathcal{L}_{\text{total}} = \mathcal{L}_{\text{distill}} + \lambda \mathcal{L}_{\text{sparsity}}$} \\
\adjustbox{max width=\linewidth}{$\mathcal{L}_{\text{sparsity}} = \frac{1}{L \cdot H \cdot T}\sum_{l,h,t} \big( g_{l,h,t} + g_{l,h,t}(1 - g_{l,h,t}) \big)$}
\end{gather*}
where $\mathcal{L}_{\text{distill}}$ is the mean squared error of the last-layer hidden states compared to the original full-attention model, and $\lambda$ serves as a hyperparameter controlling the tradeoff between representation fidelity and KV sparsity. Specifically, the sparsity loss $\mathcal{L}_{\text{sparsity}}$ incorporates two regularization terms for $g_{l,h,t}$: the first term induces sparsity by minimizing global cache admission, while the second penalizes non-binary values, encouraging the gate to converge towards discrete decisions (0 or 1). During inference, we apply a hard threshold $\tau$ to binarize $g$ via $\mathbb{1}(g \ge \tau)$ for admission decisions.

\section{System Implementation}
\label{sec:implementation}

Implementing WG-KV presents a unique systems challenge: unlike standard Transformers where the KV cache grows uniformly across all heads, our admission policy results in a \textit{ragged cache structure} with highly irregular sequence lengths across heads (as discussed in \cref{subsec:ragged}). To translate theoretical sparsity into wall-clock speedup and memory reduction, we present a \textbf{hardware-aware implementation} that leverages paged memory management to handle this irregular cache growth efficiently.

\myFloat{management}{t}{0.99}

\subsection{Dual-cache paged memory management}
\label{subsec:dual}

To address memory fragmentation caused by head-specific admission, we decouple the \textit{logical view} of the cache from its \textit{physical storage}. We partition the logical KV cache of each attention head into the \textbf{local cache} and \textbf{global cache} (\cref{fig:management}a), backed by a unified physical \textbf{KV pool} (\cref{fig:management}b). To bridge these views, we maintain a per-head \textbf{page table} (\cref{fig:management}c) that maps the logical pages to non-contiguous physical pages (16 tokens per page). This allows the global cache to grow dynamically without requiring contiguous reallocation.

\subsection{Efficient prefilling via sparse attention}
\label{subsec:prefill}

In the prefill phase, the system processes the entire input sequence in parallel. The goal is to compute attention and populate the KV cache efficiently.

\paragraph{Vertical-slash sparse attention.}
We leverage the sparsity pattern dictated by score $g$ to accelerate attention computation. Following the Write-Gated Attention formulation, the attention mask $M$ exhibits a ``vertical-slash'' pattern (\cref{fig:overview}b): tokens attend to preceding \textit{globally} high-utility tokens (``vertical'') and their \textit{local} neighborhood (``slash''). Formally, for query $i$, key $j$, $M_{ij}$ is defined as:
\begin{myEquation}
    M_{ij} = \big(\mathbb{1}(i - j < W_{\text{local}}) \lor \mathbb{1}(g_j \ge \tau)\big) \land \mathbb{1}(i \geq j)
\end{myEquation}

We utilize the sparse FlashAttention kernel from MInference \cite{minference} to compute attention efficiently. This avoids the cost of dense $O(N^2)$ attention by skipping masked-out regions.

\paragraph{Initial cache population.} Concurrently with attention computation, we write KV states to memory. Tokens within the final $W_{\text{local}}$ window are written to the local cache. Tokens preceding this window are written to the global cache only if their $g \ge \tau$; otherwise, they are discarded immediately.

\subsection{Efficient decoding via lazy promotion}

In the decode phase, tokens are generated autoregressively. Unlike standard inference systems \cite{pagedattn,sglang} that continuously append new KVs to a uniformly growing cache, WG-KV manages the head-wise irregular cache lengths efficiently via a \textbf{lazy promotion} strategy. This mechanism enforces the local window size $W_{\text{local}}$ via a ring buffer while selectively persisting high-utility states.

As illustrated in \cref{fig:management}d, for a given head, upon generating a new token $(k_t, v_t)$ with score $g_t$, the system (1) inspects the ``victim'' token at the current \textbf{local pointer}. If the victim's stored score satisfies $g \ge \tau$, it is (2) promoted to the global cache, advancing the \textbf{global pointer}. The new token then (3) overwrites the victim in the local cache, and the local pointer is incremented modulo $W_{\text{local}}$. To compute attention over the resulting ragged cache, we can fold the head dimension into the batch dimension to leverage variable-length PagedAttention \cite{pagedattn} kernel, with details provided in Appendix~\ref{app:kernel}.
\section{Experiments}

\myFloat{benchmark}{t}{1}

In this section, we evaluate WG-KV across (1) the sparsity-accuracy tradeoff and (2) practical system efficiency, followed by (3) a qualitative analysis of its admission patterns and additional analyses.

\subsection{Experimental setup}

\paragraph{Model and training.} We implement WG-KV on top of Llama-3.1-8B \cite{llama3} and Qwen3-4B-2507 \cite{qwen3}. We freeze the backbone model and train only the lightweight Write-Gate MLP. Training is performed on a subset of FineWeb-Edu \cite{fineweb} with the objective of minimizing  $\mathcal{L}_{\text{total}}$. Detailed training configurations are provided in Appendix~\ref{app:training}.

\paragraph{Evaluation scope.} We analyze the results on Llama-3.1-8B in this section. To demonstrate the generalizability of WG-KV across different model architectures, we provide additional evaluation results on Qwen3-4B-2507 in Appendix~\ref{app:qwen}.

\subsection{Sparsity-accuracy tradeoff}
\label{sec:tradeoff}

\paragraph{Benchmarks.} To evaluate the robustness of our learnable admission policy, we utilize HELMET \cite{helmet}, a comprehensive benchmark for long-context understanding. Our evaluation covers 14 tasks, spanning five distinct categories: \textit{Retrieval Augmented Generation}, \textit{Passage Reranking}, \textit{Long-Document QA}, \textit{Summarization}, and \textit{Many-Shot In-Context Learning}. Detailed benchmark descriptions are provided in Appendix~\ref{app:benchmark}.

\paragraph{Baselines.} We compare against three representative baselines which we re-contextualize as static (input-agnostic) or training-free admission policies. This setup ensures a fair comparison where all baselines can leverage the system-level efficiency gains of KV Admission as described in \cref{sec:implementation}. We evaluate \textbf{Local Attention}, which employs a uniform static policy retaining only recent and initial sink tokens \cite{streamingllm}; \textbf{DuoAttention} \cite{duoattn}, a head-wise static policy that designates specific heads as retrieval or streaming heads based on pre-computed profiles; and \textbf{AdaEA++}, a training-free admission policy adapted from Expected Attention \cite{expectedattn} and AdaKV \cite{adakv}. Detailed baseline descriptions are provided in Appendix~\ref{app:baseline}.

\paragraph{Setup.} We investigate the efficacy of different admission policies in preserving task accuracy under varying levels of KV sparsity. For WG-KV, we fix the binarization threshold $\tau=0.1$ while sweeping $\lambda$ to control the admission budget to achieve different sparsity targets. For baselines, we vary the sliding window size (Local Attention), the ratio of retrieval heads (DuoAttention), or the score threshold (AdaEA++). \cref{fig:benchmark} shows the results.

\paragraph{Results.} WG-KV outperforms baselines across almost all sparsity configurations, with the most prominent advantages under tight admission budgets (e.g., admitting $<$40\% KVs). Notably, in information-intensive tasks such as \textit{Passage Reranking} (MS MARCO) and \textit{Summarization} (InfiniteBench Sum and Multi-LexSum), WG-KV maintains near-lossless task accuracy even when operating at 90\% sparsity (admitting only 10\% KVs). In contrast, Local Attention degrades rapidly as the admission budget tightens; DuoAttention and AdaEA++, while more robust, lack the adaptability to achieve parity with WG-KV. Their reliance on static head profiles or training-free attention statistics prevents them from capturing complex context dependencies, resulting in severe accuracy drops in high-sparsity regimes.

\subsection{System efficiency}
\label{sec:system_performance}

\myFloat{perf}{b}{1}

\paragraph{Setup.} Having established WG-KV's ability to maintain near-lossless accuracy at extreme sparsity, we validate its practical system efficiency under realistic workloads. We measure the end-to-end latency and memory usage using input sequences sampled from the PG-19 \cite{pg19} long-document dataset. We evaluate the model equipped with WG-KV at $\lambda = 0.16$, which achieves approximately 80\% sparsity (i.e., admitting 20\% KVs), and benchmark it against the standard full-attention model (see Appendix~\ref{app:system} for detailed experimental setup). \cref{fig:perf} shows the results.

\paragraph{Results.} Our efficient hardware-aware implementation (\cref{sec:implementation}) effectively translates theoretical sparsity into practical system gains. In the \textit{prefill} phase, WG-KV achieves a \textbf{2.56--3.85x} speedup across varying sequence lengths, primarily attributed to the vertical-slash sparse attention mechanism, which bypasses the massive computational cost of dense full attention. In the \textit{decode} phase, it achieves a \textbf{1.61--2.63x} speedup as the reduced KV cache size alleviates memory transfer overhead and consequently lowers attention latency. This compact KV cache also translates to a \textbf{36--60\%} reduction in overall memory usage.

\paragraph{Overhead analysis.} The introduction of the Write-Gate MLP incurs minimal overhead. In terms of model size, the additional parameters constitute only $\approx$0.4\% of the total parameter count in both Llama and Qwen models. In terms of computation, as indicated by the slight increase in non-attention latency (which includes the execution time of the MLP) in \cref{fig:perf}a-b, the introduced overhead is negligible. The massive reduction in attention computation and memory consumption far outweighs the marginal cost of the MLP.

\subsection{Qualitative analysis of admission patterns}

To explore the internal mechanics of WG-KV, we visualize the admission decisions across different attention heads in \cref{fig:interpret_short}, using an excerpt from \citet{medtner} as the input text. Remarkably, without any explicit rules, WG-KV autonomously learns highly specialized, head-specific memory allocation strategies. We observe a clear division of labor: certain heads act as structural and syntactic anchors by predominantly admitting function words (\cref{fig:interpret_short}a), punctuation marks (\cref{fig:interpret_short}b), or newline characters (\cref{fig:interpret_short}d). Conversely, other heads selectively persist information-dense tokens such as proper nouns (\cref{fig:interpret_short}c) and numerical values (\cref{fig:interpret_short}e). Some heads exhibit complex, non-trivial admission behaviors (\cref{fig:interpret_short}f-g) that defy simple lexical categorization, possibly tracking high-level semantic dependencies. This emergent specialization demonstrates WG-KV's capability to intelligently compose a compact yet comprehensive global cache. Full visualizations are provided in Appendix~\ref{app:interpret}.

\subsection{Additional analyses}

To further investigate WG-KV's characteristics, we provide additional analyses. (1) Appendix~\ref{app:synergy} explores the synergy between WG-KV and Selection/Eviction methods. (2) Appendix~\ref{app:hyperparam} analyzes hyperparameters $\lambda$ and $\tau$, and compares continuous with discrete gating. (3) Appendix~\ref{app:local} ablates the local cache to validate its necessity. (4) Appendix~\ref{app:dynamic} demonstrates WG-KV's input-dependent admission patterns and its adaptive memory allocation across task-specific semantic structures.

\myFloat{interpret_short}{t!}{0.95}
\section{Conclusion}

In this paper, we address the inefficiency of long-context LLM inference by formalizing KV Admission as a critical third primitive alongside KV Selection and Eviction. We introduce Write-Gated KV (WG-KV), a learnable mechanism that predictively filters redundant states pre-write, achieving substantial efficiency gains while preserving near-lossless task accuracy. Backed by our hardware-aware implementation and extensive evaluations, we demonstrate that \textit{learning what to write} provides a principled and practical recipe for efficient long-context inference.
\section*{Limitations}

WG-KV is a first step toward learnable KV Admission, but several limitations remain: 
(1) Our evaluation is restricted to dense models in the 4B--8B parameter range; generalizability to larger scales, Mixture-of-Experts architectures, multimodal LLMs, and quantized KV cache \cite{kvquant} remains an open question.
(2) WG-KV uses a simple two-layer Write-Gate MLP trained on top of a frozen backbone. More expressive gate architectures or fully integrated training that internalizes admission behavior into model weights may yield better sparsity-quality tradeoffs, which we leave for future exploration.
(3) While WG-KV is evaluated extensively on HELMET, our study does not yet cover multilingual settings \cite{oneruler}, multi-turn dialogues \cite{multichallenge}, long-horizon agentic workflows \cite{worfbench}, or ultra-long context regimes scaling up to millions of tokens \cite{longcodebench}. These settings may induce different token-utility patterns and require dedicated large-scale evaluation.
(4) Our evaluation in \cref{sec:tradeoff} is limited to benchmark accuracy; open-ended generation quality---such as hallucination, faithfulness, long-form coherence, human preference, and response safety---is not explicitly evaluated. Validating WG-KV against these metrics remains an important direction for future research.
(5) On the systems side, our efficiency results are limited to a single-batch, single-GPU setup. Realizing these speedups in complex production environments---such as multi-GPU serving \cite{helix}, prefix caching \cite{kvflow}, and speculative decoding \cite{specinfer}---introduces distinct engineering challenges that warrant further system-level exploration.
\section*{Ethical Considerations}

Because WG-KV filters tokens from the KV cache, there is a risk that critical safety guardrails, alignment instructions, or system prompts could be inadvertently discarded over extended contexts. Adversaries could potentially exploit this mechanism by crafting malicious inputs designed to manipulate the gating scores and intentionally flush safety-critical tokens, thereby facilitating jailbreak attacks \cite{attack}. Future deployments should incorporate targeted mitigation strategies to ensure that efficiency gains do not compromise model safety and alignment.
\section*{Acknowledgments}

This work was supported in part by the National Science and Technology Council (NSTC), Taiwan, under Grants NSTC 113-2628-E-001-004-MY3, NSTC 114-2221-E-002-180-MY3, and NSTC 115-2223-E-002-001-MY3; Academia Sinica under Grant AS-IA-113-M04-ASSA; and the Center of Data Intelligence: Technologies, Applications, and Systems at National Taiwan University under Grant 115L9009, as part of the Featured Areas Research Center Program within the Higher Education Sprout Project funded by the Ministry of Education (MOE), Taiwan. We also thank Shih-Hsuan Chiu for valuable discussions on this work, and the National Center for High-performance Computing (NCHC), Taiwan, for providing computational resources.

\bibliography{includes/references}

\appendix

\section{Extended Related Work}
\label{app:related}

\paragraph{Static and head-level attention specialization.} Studies on attention dynamics reveal that token utility is structurally skewed across heads and positions. \citet{streamingllm} identify ``attention sinks,'' demonstrating that retaining a few initial tokens allows for stable streaming inference without the full KV cache. Building on this, \citet{fastgen} propose profiling attention heads to statically categorize them for distinct retention policies. Mechanistic studies \cite{retrievalhead} further reveal that specific ``retrieval heads'' are uniquely critical for long-range dependencies. Leveraging these characteristics, \citet{razorattn}, \citet{duoattn}, and \citet{prulong} enhance inference efficiency by statically identifying retrieval-critical heads to retain full context, while restricting others to local context. Unlike these static heuristics, our approach introduces a learnable, input-dependent admission mechanism that captures dynamic token utility patterns at runtime.

\paragraph{Read-time KV Selection.} A large body of work accelerates long-context inference by selectively attending to a sparse, query-dependent subset of cached KV pairs at each step. These methods approximate attention while maintaining the complete underlying cache for future queries. Existing approaches exploit sparsity in various dimensions, including temporal \cite{hip,specache,attnpredictor}, channel \cite{sparq}, inter-layer \cite{omnikv,hshare}, head-wise \cite{moh,qada}, top-$p$ \cite{twilight}, and low-rank \cite{sea,loki} structures. Some retrieval-oriented designs treat the KV cache as an indexed memory \cite{arkvale,retrievalattn,squeezedattn,pqcache,clusterkv,magicpig,retroinfer}. Systems-oriented designs further integrate selection with KV-aware paging \cite{quest,lserve}, runtime management \cite{infinigen,shadowkv}, sparse prefill patterns \cite{minference,flexprefill,spargeattn,xattn}, or learnable indexers \cite{seerattn,moba,nsa,seerattnr,deepseekv32,deepseekv4}. Our write-time admission mechanism is complementary to these read-time strategies, allowing for compound efficiency gains by filtering redundant states prior to selection.

\paragraph{Post-write KV Eviction.} Another line of work focuses on maintaining a bounded cache by retrospectively pruning previously accumulated states. Representative methods employ statistics derived from past attention \cite{scissorhands,h2o,tova} or intrinsic token properties \cite{sirllm,l2norm,qfilters,expectedattn} to evict tokens. Other approaches refine the eviction criteria through learnable interaction scores \cite{dcp,attngate}, importance estimation \cite{locret,kvzap,kvp,trimkv,fastkvzip}, prompt-guided heuristics \cite{snapkv,nacl,pyramidinfer,finch,andpro}, reconstruction objectives \cite{kvzip}, or reasoning trace redundancy \cite{dms,headkv,rkv}. Unlike Admission, which pre-filters redundant states at the source, Eviction manages cache bounds by removing obsolete history; thus, the two serve distinct, complementary roles in the token lifecycle.

\paragraph{Representation-level KV compression.} A parallel line of research seeks to compress the KV representations themselves. Instead of discarding tokens, merging-based methods aggregate information by combining keys and values based on semantic similarity \cite{kvmerger}, weighted accumulation \cite{cam}, low-rank residual recovery \cite{less}, or learnable mechanisms \cite{dmc,deepseekv4}. Additionally, dimensionality reduction techniques exploit inherent redundancy in hidden states by projecting KV pairs into lower-rank subspaces \cite{eigenattn,shadowkv,palu,think,matryoshkakv}, leveraging inter-layer redundancy \cite{minicache,lckv,cla,xkv}, or applying quantization \cite{kvquant}. These lines of work are largely orthogonal to our admission mechanism.
\section{PagedAttention Kernel Compatibility}
\label{app:kernel}

Our head-specific admission policy results in a ragged KV cache structure where sequence lengths vary significantly across heads. To perform attention operations over this irregular structure, we can employ vLLM's PagedAttention \cite{pagedattn} kernel, which natively supports variable sequence lengths across the batch dimension. By reshaping the attention input---folding the head dimension ($H$) into the batch dimension ($B$)---the computation can be treated as $B \times H$ independent sequences of varying lengths. This allows existing serving systems to apply industry-standard CUDA kernels to compute attention over the ragged KV cache seamlessly.
\section{Additional Details for Training}
\label{app:training}

Training is performed on a subset of FineWeb-Edu \cite{fineweb}, using samples with lengths ranging from 4K to 32K. The training consists of 7,500 samples with a batch size of 1, totaling approximately 63M tokens (merely 0.00042\% of Llama-3.1-8B's $\approx$15T pre-training tokens). Each training run requires approximately 6 hours on a single H100 GPU. We employ the AdamW optimizer \cite{adamw} with a weight decay of 0.01 and a cosine learning rate schedule (linear warmup for the first 10\% of steps) peaking at 0.001. To simulate instruction-following scenarios, we prepend a generic prompt (``\textit{Please analyze and summarize the following text:\textbackslash n\textbackslash n}'') to each sample. 

For reasoning models, we augment the training set with an additional 7,500 samples from Nemotron-Math-v2 \cite{nemotronmath}. While this addition doubles the training cost, we empirically observe that the resulting model consistently outperforms the version trained without this augmentation on reasoning-intensive benchmarks.
\section{Additional Details for HELMET Benchmark}
\label{app:benchmark}

We evaluate our method on 14 tasks from the HELMET benchmark \cite{helmet}. The tasks are categorized as follows:
\begin{itemize}
    \item \textbf{Retrieval augmented generation:} Evaluated on Natural Questions \cite{nq}, TriviaQA \cite{triviaqa}, PopQA \cite{popqa}, and HotpotQA \cite{hotpotqa}.
    \item \textbf{Passage reranking:} Evaluated on MS MARCO \cite{msmacro}.
    \item \textbf{Long-document QA:} Evaluated on NarrativeQA \cite{narrativeqa}, InfiniteBench QA \cite{infinitebench}, and InfiniteBench MC \cite{infinitebench}.
    \item \textbf{Summarization:} Evaluated on InfiniteBench Sum \cite{infinitebench} and Multi-LexSum \cite{lexsum}.
    \item \textbf{Many-shot in-context learning:} Evaluated on TREC Fine \cite{trec}, NLU \cite{nlu}, BANKING77 \cite{banking}, and CLINC150 \cite{clinc}.
\end{itemize}
All tasks are evaluated with 32K context length using the official evaluation scripts provided by HELMET.
\section{Additional Details for Admission Baselines}
\label{app:baseline}

We compare our method against static (input-agnostic) or training-free admission baselines. The specific configurations for each baseline are detailed below.

\paragraph{Local Attention.} A uniform static admission policy that only admits the most recent tokens while retaining a small set of initial tokens \cite{streamingllm} to serve as attention sinks.

\paragraph{DuoAttention \cite{duoattn}.} A head-wise static policy that categorizes attention heads into retrieval heads (which retain the full KV cache) and streaming heads (which retain the local sliding window and initial attention sinks). We utilize the official profiles provided by the authors for the Llama-3.1-8B model. For Qwen3-4B-2507, as no official profile is available, we apply a similar optimization-based identification method described in the paper to determine the head allocation.

\paragraph{AdaEA++.} Expected Attention (EA, \citealp{expectedattn}) and AdaKV \cite{adakv} are originally designed as post-write KV Eviction methods. To fairly benchmark them under our KV Admission framework, we adapt them into AdaEA++, a pre-write training-free admission policy. Using the same FineWeb-Edu subset utilized for training WG-KV (Appendix~\ref{app:training}), we first precompute the mean and covariance of queries for each head offline. We then apply EA's closed-form scoring on these samples and adopt AdaKV's head-wise allocation strategy to find the top-$k$ score thresholds for a target admission budget.  During inference, AdaEA++ calculates the expected attention score for each newly generated KV pair using the precomputed query statistics. If the score surpasses the threshold, the token is admitted. This adaptation allows AdaEA++ to function as a strong training-free admission baseline that is fully compatible with our efficient system implementation (\cref{sec:implementation}).

\myParagraphWithoutTitle To ensure a fair comparison, we standardize the cache retention configurations across all baselines. Specifically, we fix the attention sink size to 128 for Local Attention and DuoAttention, and set the local sliding window size to 256 for DuoAttention and AdaKV++ to align with WG-KV. 
\section{Detailed Experimental Setup for System Efficiency}
\label{app:system}

\myFloat{benchmark_qwen}{t}{1}

\myFloat{perf_qwen}{t}{1}

This section details the software stack, attention kernels, and evaluation setup underpinning the latency results reported in \cref{sec:system_performance}. By rigorously accounting for all runtime overheads, we demonstrate that our results reflect the real execution time of a fully functional system prototype rather than theoretical estimates.

\subsection{Software stack and attention kernels}
To ensure a fair comparison, both the full-attention baseline and WG-KV are built upon Hugging Face Transformers \cite{transformers}. We utilize \texttt{AutoModelForCausalLM} for inference but replace the core attention kernels as follows:

\paragraph{Prefill phase.} We employ vLLM's \texttt{sparse\_attn\_func} kernel, which is an efficient sparse FlashAttention kernel based on MInference \cite{minference}. The full-attention baseline operates with a standard causal mask, whereas WG-KV utilizes the kernel to enforce the vertical-slash sparse attention pattern.

\paragraph{Decode phase.} We employ a custom Triton-based \cite{triton} attention kernel. The full-attention baseline operates with a standard KV cache, whereas WG-KV utilizes the kernel to handle the ragged KV cache structure. Although this ragged structure can be mapped to vLLM's PagedAttention \cite{pagedattn} kernel via batch-folding (as detailed in Appendix~\ref{app:kernel}), we empirically observed that extreme variance in sequence lengths across the folded batch introduces workload imbalance, which hinders optimal GPU utilization. Our custom kernel mitigates this by dynamically balancing the workload across heads via a split-reduce strategy \cite{flashdecoding}, which redistributes the ragged KV blocks across all heads into uniform-sized chunks to maximize hardware efficiency.

\myParagraphWithoutTitle By standardizing the underlying attention kernels, we ensure that the observed speedups are attributable solely to algorithmic efficiency rather than differences in kernel optimizations.

\subsection{System implementation and overhead accounting}
Our evaluations measure the performance of a fully implemented dual-cache paged memory system. The reported latency includes the elapsed time for all CPU-side and GPU-side operations required to handle the admission decisions and the resulting ragged KV cache structure. Specifically, the timing loop encompasses the forward computation of the Write-Gate MLP, the allocation of the KV cache, the management of the page table mapping, and the generation of block indices required by the \texttt{sparse\_attn\_func} kernel. During the decoding phase, the measurement also includes the full overhead of the lazy promotion logic. This involves the operations to check the victim token, update the page tables, and migrate tokens from local to global cache when necessary. This comprehensive accounting ensures that the reported speedups faithfully translate to real-world efficiency.

\subsection{Evaluation setup}
To accurately reflect realistic workloads, evaluations are performed using input sequences sampled from the PG-19 \cite{pg19} long-document dataset. All experiments are conducted on a single H200 GPU with a batch size of 1. Latency is recorded via CUDA Events after one warm-up pass. We report the end-to-end Time to First Token (TTFT) for the prefill phase, and the average Time per Output Token (TPOT) over 300 generation steps for the decode phase.
\section{Evaluation Results on Qwen Model}
\label{app:qwen}

\myFloat{quest}{t}{1}

\myFloat{integration_new}{t}{1}

To demonstrate the generalizability of WG-KV across different model architectures, we perform additional evaluations on the non-reasoning variant of Qwen3-4B-2507 \cite{qwen3}.

\subsection{Sparsity-accuracy tradeoff}

\cref{fig:benchmark_qwen} shows the evaluation results on the HELMET benchmark. Similar to the observations on Llama-3.1-8B (\cref{fig:benchmark}), WG-KV maintains high accuracy on Qwen3-4B-2507 across varying admission budgets, consistently outperforming baselines across diverse tasks.

\subsection{System efficiency}

Following the same experimental setup used for Llama-3.1-8B, we measure the end-to-end latency and memory usage on Qwen3-4B-2507 at $\lambda=0.32$, which achieves approximately 80\% sparsity (i.e., admitting 20\% KVs). As shown in \cref{fig:perf_qwen}, WG-KV achieves a \textbf{3.23--4.17x} speedup in prefill and a \textbf{1.59--2.50x} speedup in decode across varying sequence lengths. Furthermore, WG-KV reduces memory usage by \textbf{53--69\%} compared to the standard full-attention model.
\section{Synergy with Selection and Eviction}
\label{app:synergy}

As conceptualized in \cref{subsec:framework}, the three KV management primitives operate at distinct stages of the token lifecycle and are inherently complementary. In this section, we demonstrate that WG-KV can be seamlessly integrated with both read-time KV Selection and post-write KV Eviction methods to achieve compound gains.

\subsection{Synergy with KV Selection} 

We investigate the interplay between WG-KV and Quest \cite{quest}, a representative KV Selection method, using Llama-3.1-8B\footnote{In this evaluation, WG-KV operates with $\lambda = 0.16$, admitting $\approx20\%$ KVs.}. During decoding, Quest selects relevant KVs by utilizing key landmarks\footnote{For Quest, ``key landmarks'' refer to the per-page minimum and maximum key vectors.} to estimate the criticality of each KV page. When evaluated at a selection budget of 2048 KVs per query, WG-KV effectively pre-filters redundant tokens, reducing KV memory usage (\cref{fig:quest}a). This compacted cache then shrinks the candidate pool for selection, decreasing memory access per decode step (\cref{fig:quest}b), which further translates into lower attention latency (\cref{fig:quest}c).

Notably, these compound efficiency gains do not come at the expense of model capability. As shown in \cref{fig:quest}d, integrating WG-KV achieves nearly identical HELMET accuracy to standalone Quest, indicating that performing selection on a cache already filtered by WG-KV preserves representation fidelity. \cref{fig:integration_new} further supports these findings by showing that this robustness is consistently maintained across varying selection budgets, highlighting the composable nature of pre-write Admission and read-time Selection.

\subsection{Synergy with KV Eviction} 
\label{subsec:evict_synergy}

\cref{fig:evict} conceptually illustrates the synergy between KV Admission and Eviction. Without Admission, rapid KV accumulation leads to steep cache growth, causing the system to frequently hit the eviction threshold and trigger aggressive evictions (\cref{fig:evict}a). By pre-filtering redundant tokens, Admission flattens the growth, delaying the onset of eviction and reducing its frequency (\cref{fig:evict}b). The smaller shaded area under this flattened curve also translates to a lower cumulative attention cost over the generation process.

\myFloat{evict}{t}{0.6}

\myFloat{snapkv_short}{t}{0.75}

We empirically validate this synergy by coupling WG-KV with SnapKV \cite{snapkv}, a representative KV Eviction method, in reasoning-intensive scenarios where long thinking traces rapidly consume KV cache memory. We evaluate this on the AIME25 benchmark \cite{aime25} using the reasoning variant of Qwen3-4B-2507 under strict memory limits. As illustrated in \cref{fig:snapkv_short}, relying solely on SnapKV (Eviction) leads to catastrophic degradation (26.7\% accuracy); this occurs because the ``write-then-throw'' inefficiency allows noise to flood the cache, triggering frequent and premature evictions that inadvertently discard critical context. Conversely, relying solely on WG-KV (Admission) to satisfy strict memory limits forces the policy to be overly aggressive, causing the model to reject potentially useful information and ``starve'' itself. However, combining these primitives yields a superior tradeoff. WG-KV acts as a gatekeeper that filters noise at the source, effectively reducing the frequency of eviction triggers and allowing the eviction policy to focus on pruning \textit{obsolete} history. This synergy stabilizes the reasoning process and restores accuracy to 80.0\%, matching the performance of the unbounded KV baseline while strictly adhering to memory limits. Detailed experimental setup and analysis are provided in Appendix~\ref{app:evict}.
\section{Details of Eviction Synergy for Long-Chain Reasoning}
\label{app:evict}

This section details the experimental setup and further analysis of the Admission-Eviction synergy introduced in Appendix~\ref{subsec:evict_synergy}, particularly in reasoning-intensive scenarios where intermediate ``thinking tokens'' cause rapid cache expansion. While WG-KV effectively filters redundant tokens at the source, real-world hardware imposes strict limits on memory capacity. This constraint is particularly pronounced in modern high-throughput serving systems \cite{pagedattn,sglang}, where multiple concurrent requests compete for a shared memory pool. To sustain inference over prolonged generation phases---such as the extended reasoning trajectories seen in complex mathematical problem-solving---the system must enforce a hard memory budget by retrospectively pruning obsolete history. Here, we demonstrate how coupling pre-write Admission with post-write Eviction effectively reserves this limited capacity exclusively for critical context, yielding a superior memory-accuracy tradeoff compared to utilizing either primitive in isolation.

\subsection{Eviction policy implementation}

To evaluate the synergy between KV Admission and Eviction, we integrate WG-KV with SnapKV to manage the cache under strict memory constraints by enforcing a hard KV cache budget such that the average capacity per head is 4096 tokens. When the cache size exceeds this budget, an eviction process is triggered to remove the bottom 10\% of KV pairs with the lowest importance scores.

The eviction policy operates independently for each KV head. Consider a specific head with cached keys $K \in \mathbb{R}^{N \times d}$ and a corresponding set of query heads $\mathcal{G}$ (as in Grouped-Query Attention). We define the importance score $S_j$ for the $j$-th key-value pair based on queries from the most recent window of size $W_{\text{obs}}=256$, denoted as $\{Q_{\text{obs}}^{(h)} \in \mathbb{R}^{W_{\text{obs}} \times d} \mid h \in \mathcal{G}\}$. The scoring process proceeds in three steps:
\begin{enumerate}
    \item \textbf{Attention computation:} We first compute the post-softmax attention scores $A^{(h)} \in \mathbb{R}^{W_{\text{obs}} \times N}$ for each query head $h \in \mathcal{G}$ against the keys $K$:
    \begin{myEquation}
    A^{(h)} = \operatorname{Softmax}\left(Q_{\text{obs}}^{(h)} K^\top/\sqrt{d}\right)
    \end{myEquation}
    \item \textbf{Score aggregation:} To capture the utility of a key across the observation window, we aggregate the scores by maximizing over the query heads in $\mathcal{G}$ and summing over $W_{\text{obs}}$:
    \begin{myEquation}
        S^{\text{raw}}_j = \sum_{i=1}^{W_{\text{obs}}} \max_{h \in \mathcal{G}} A_{i,j}^{(h)}
    \end{myEquation}
    \item \textbf{Local smoothing:} Finally, we apply a max-pooling operation with a kernel size of $W_{\text{pool}}=5$ over the sequence length $N$:
    \begin{myEquation}
        S = \operatorname{MaxPool}\left(S^{\text{raw}},W_{\text{pool}}\right)
    \end{myEquation}
\end{enumerate}

\subsection{Quantitative analysis}

We evaluate the impact of WG-KV on the AIME25 benchmark using the reasoning variant of Qwen3-4B-2507. We analyze the system's behavior in two distinct scenarios: (a) an unbounded memory setting to isolate the behavior of WG-KV, and (b) a bounded memory setting to analyze its synergy with SnapKV.

\myFloat{snapkv}{t}{0.7}

\paragraph{Admission alone is insufficient under strict bounds.}
\cref{fig:snapkv}a illustrates the tradeoff when no size limit is enforced on the KV cache. Increasing $\lambda$ reduces the KV cache size but progressively degrades accuracy.
In the bounded setting (\cref{fig:snapkv}b), relying solely on Admission to avoid hitting the memory limit necessitates an overly aggressive policy. For instance, with $\lambda=1.28$, the admission rate is low enough that the eviction trigger count drops to zero (the cache never fills up). However, this comes at a steep cost: accuracy drops to 66.7\%. This indicates that when using Admission alone, the model is forced to reject tokens that are useful, starving itself of information to satisfy the memory constraint.

\paragraph{Eviction alone leads to catastrophic degradation.}
Conversely, relying solely on Eviction (the ``Off'' baseline in \cref{fig:snapkv}b) results in a catastrophic accuracy drop. Without WG-KV, the cache grows rapidly with noise tokens, triggering frequent and premature evictions that inadvertently discard critical information. Consequently, accuracy collapses to 26.7\%.

\paragraph{Optimal performance requires combining Admission and Eviction.}
The synergy between the two primitives achieves the best of both worlds. By setting a moderate admission budget ($\lambda=0.32$), WG-KV prevents redundant tokens from being admitted into the global cache. This filtering process allows the eviction policy to focus on its primary strength: identifying and removing \textit{obsolete} information that has outlived its usefulness. As shown in \cref{fig:snapkv}b, pairing WG-KV with SnapKV allows the model to achieve 80\% accuracy under strict memory constraints---matching the performance of the original model with unbounded memory as shown in \cref{fig:snapkv}a (``Off''). This result confirms that Admission and Eviction are not mutually exclusive but complementary strategies essential for memory-constrained long-chain reasoning.

\subsection{Qualitative analysis}

To qualitatively illustrate the synergy between KV Admission and Eviction, \cref{fig:evict_interpret} visualizes the token lifecycle during an AIME25 reasoning trace. The visualization reveals an intriguing phenomenon: fundamental semantic anchors, such as the initial problem statement (\cref{fig:evict_interpret}a), critical sub-conclusions (\cref{fig:evict_interpret}c), and the final definitive answer validation (\cref{fig:evict_interpret}e), are admitted and persistently retained to guide the global reasoning trajectory and formulate the final response. Conversely, intermediate exploratory hypotheses and trial calculations (\cref{fig:evict_interpret}b and \cref{fig:evict_interpret}d) are initially admitted by WG-KV to support immediate derivation steps, but as the model reaches intermediate milestones and transitions to subsequent logical phases, these temporary ``scratchpad'' states are retrospectively identified as obsolete and pruned by SnapKV. This demonstrates how the joint policy effectively distills convoluted thinking traces into a compact cache without sacrificing the information needed for accurate problem-solving.

\myFloat{evict_interpret}{t}{0.95}
\section{Hyperparameter Analysis}
\label{app:hyperparam}

\subsection{Impact of \texorpdfstring{$\lambda$}{λ} and \texorpdfstring{$\tau$}{τ}}

In this section, we provide a detailed analysis of the impact of the training-time hyperparameter $\lambda$ and the inference-time binarization threshold $\tau$. \cref{fig:threshold} plots the average distillation loss $\mathcal{L}_{\text{distill}}$ computed on a held-out validation set of 750 FineWeb-Edu samples against the normalized KV cache size. By sweeping $\lambda$ and $\tau$, we trace the Pareto frontier of the model's sparsity-loss tradeoff. Increasing $\lambda$ effectively drives the model towards higher sparsity (admitting less KVs), albeit with an increase in $\mathcal{L}_{\text{distill}}$. Empirically, we observe that fixing $\tau \approx 0.1$ consistently yields near-optimal operating points along this frontier across various $\lambda$ settings. We therefore adopt $\tau=0.1$ for all of our experiments.

\subsection{Continuous versus discrete gating}

\myFloat{threshold}{t}{0.78}

\myFloat{threshold_hc}{t}{0.78}

Since admission is inherently a discrete decision, binarizing a continuously relaxed gate (\cref{subsec:gate,subsec:objective}) raises the concern of a train-inference mismatch: a token with an intermediate gating score remains partially visible during training, yet is either fully retained or entirely discarded at inference. To quantify this gap, we retrain the Write-Gate MLP with hard-concrete gates \cite{hardconcrete}, reinterpreting the MLP output as an unnormalized log-probability $\log \alpha$ (temperature 0.667 with stretch interval from -0.1 to 1.1) and replacing $\mathcal{L}_{\text{sparsity}}$ with the closed-form expected $L_0$ norm. Owing to the stretch-and-clip construction, the resulting gate is exactly binary for the vast majority of tokens and is therefore nearly invariant to $\tau$ by design, serving as a mismatch-free reference point.

\cref{fig:threshold_hc} compares the two variants. Across the admission budgets adopted in this paper, the two frontiers are nearly identical, indicating that the binarization term in $\mathcal{L}_{\text{sparsity}}$ suffices to keep the learned policy close to a discrete one and that the relaxation gap is small in practice. The two diverge only under aggressive budgets, where the discrete gate traces a modestly lower frontier.

Since the two perform comparably at the budgets we target, we adopt the continuous gate, which avoids the additional gradient variance and temperature sensitivity of stochastic sampling \cite{concrete,gumbel}.

\myFloat{threshold_cmp}{t}{0.78}

\myFloat{gate}{t}{1}
\section{Ablation Study on Local Cache}
\label{app:local}

To validate the necessity of local cache within WG-KV, we conducted an ablation study by removing the local cache. Specifically, we retrained the model using the same objective described in \cref{subsec:objective}, but constrained the local window size $W_{\text{local}}$ to 1. This configuration forces the model to rely exclusively on the learnable gate to decide whether to retain any token \textit{immediately upon generation}, effectively eliminating the grace period for recent tokens.

\cref{fig:threshold_cmp} compares the full WG-KV system against the variant without local cache (``WG-KV w/o Local Cache''). We plot the average distillation loss $\mathcal{L}_{\text{distill}}$ on a held-out validation set of 750 FineWeb-Edu samples against the normalized KV cache size.

The results highlight a sharp degradation in the variant without local cache as admission budget tightens. This empirical evidence strongly substantiates the ``transient utility'' hypothesis discussed in \cref{subsec:sparsity}: recent tokens, even those with negligible long-term utility, attract dense local attention. Without a dedicated local cache to provide a temporary grace period, the admission policy is forced to make premature retention decisions for these high-saliency states. Consequently, the system fails to decouple immediate context needs from long-term retention, confirming that the dual-cache design---combining a transient local cache with a persistent global cache---is essential for efficient and robust long-context inference.
\section{Analysis of WG-KV's Input-Dependent Admission Patterns}
\label{app:dynamic}

To demonstrate that WG-KV learns an input-dependent admission policy rather than degenerating into input-agnostic head specialization, we provide both qualitative and quantitative analyses in this section.

\subsection{Qualitative analysis}
\label{app:qualitative}

To show that WG-KV adapts its memory allocation based on the input context, \cref{fig:gate} visualizes the distribution of KV cache sizes across all attention heads using Llama-3.1-8B with $\lambda=0.08$. We compare two distinct long-context tasks: \textit{(a) Code Summarization}, utilizing a Python sample from The Stack \cite{thestack}, and \textit{(b) HTML to TSV}, employing a structured data extraction sample from LongProc \cite{longproc}.

\paragraph{Observations.} The heatmaps reveal distinct retention characteristics tailored to the input task:
\begin{enumerate}
    \item \textbf{Task sensitivity.} For \textit{(a) Code Summarization}, the cache distribution is relatively uniform, indicating that understanding code dependencies requires retaining context across diverse heads. In contrast, for \textit{(b) HTML to TSV}, the pattern is significantly sparser, suggesting that the relevant information for this task is highly concentrated. The model effectively filters out most tokens, likely preserving primarily the critical structural cues (e.g., tags) to perform the extraction.
    \item \textbf{Head-specific decisions.} Consistent with \cref{fig:distribution}, the admission policy is fine-grained. Specific heads retain nearly full history, while adjacent heads may discard most tokens.
\end{enumerate}

These findings demonstrate that WG-KV adaptively modulates its memory allocation in response to the semantic structures of the task, dynamically transitioning between high-retention and high-sparsity modes.

\subsection{Quantitative analysis}

To further justify that admission decisions are dynamically driven by the input rather than relying on input-agnostic head profiles, we conduct a cross-sequence replay experiment. We randomly sample 1,000 pairs of 8K-length sequences $(S_1,S_2)$ from the PG-19 \cite{pg19} long-document dataset and feed them into Llama-3.1-8B with $\lambda=0.08$. We measure the average MSE of the last-layer hidden states for $S_1$ against the standard full-attention model under three policies:

\begin{enumerate}
    \item \textbf{Own Policy:} Using $S_1$'s own dynamically generated admission decisions.
    \item \textbf{Replay Policy:} Forcing $S_1$ to adopt the exact admission decisions generated by $S_2$.
    \item \textbf{Random Policy:} Randomly admitting tokens for $S_1$ while matching the exact retention ratio of $S_2$.
\end{enumerate}

If WG-KV merely degenerated into learning input-agnostic head specialization, the \textit{Replay Policy} should yield an MSE close to \textit{Own Policy} and outperform \textit{Random Policy}. However, as shown in \cref{tab:replay}, the \textit{Replay Policy} performs almost identically to the \textit{Random Policy} and is significantly worse than \textit{Own Policy}. This stark contrast quantitatively confirms that WG-KV's admission decisions are strictly input-dependent and dynamically tailored to the context of the given sequence.

\myFloat{replay}{t}{0.9}
\section{Full Visualizations of WG-KV's Admission Patterns}
\label{app:interpret}

In this section, we provide the full visualizations of WG-KV's admission patterns on Llama-3.1-8B at $\lambda=0.16$. 

\crefrange{fig:l0h7}{fig:l18h6} present the full-text admission patterns corresponding to the cropped snippets shown in \cref{fig:interpret_short}a-g. These visualizations demonstrate the fine-grained division of labor among individual attention heads, ranging from lexical and structural anchors to those capturing complex semantic dependencies.

\cref{fig:interpret_summary} further shows the aggregated admission decisions across all attention heads. The color intensity corresponds to the proportion of heads that admit a given token into their global cache. We observe a strong cross-head consensus for specific token categories: special control tokens (e.g., \texttt{<|begin\_of\_text|>}), key entities, proper nouns, and semantically dense words are highly admitted. Conversely, function words and pronouns are rarely admitted. This emergent behavior, interestingly, closely mirrors human reading cognition, where semantic anchors are committed to memory while syntactic fillers are glossed over---a strategy the model learns entirely without explicit linguistic supervision.

\myFloat{l0h7}{t}{0.95}

\myFloat{l4h7}{t}{0.95}

\myFloat{l24h0}{t}{0.95}

\myFloat{l27h2}{t}{0.95}

\myFloat{l28h7}{t}{0.95}

\myFloat{l13h1}{t}{0.95}

\myFloat{l18h6}{t}{0.95}

\myFloat{interpret_summary}{t}{0.95}
\section{Artifacts and Intended Use}

Our work utilizes publicly available models, including Llama-3.1-8B \cite{llama3}, Qwen3-4B-2507 \cite{qwen3}; datasets, including PG-19 \cite{pg19}, The Stack \cite{thestack}, FineWeb-Edu \cite{fineweb}, Nemotron-Math-v2 \cite{nemotronmath}, HELMET \cite{helmet}, AIME25 \cite{aime25}, LongProc \cite{longproc}; software, including Triton \cite{triton}, Hugging Face Transformers \cite{transformers}, FlashAttention \cite{flashattn}, vLLM \cite{pagedattn}, LM Evaluation Harness \cite{lmeval}, MInference \cite{minference}, DuoAttention \cite{duoattn}, FlexAttention \cite{flexattn}, KVPress \cite{expectedattn}. Our use of these artifacts is strictly for academic research and complies with their respective open-source licenses and intended uses.
\fi

\end{document}